%% file: Main.tex
\pdfoutput=1
\documentclass[11pt,phd,a4paper,oneside]{ucl_thesis}

\input{MainPackages}

\input{Figures}

\begin{document}

\include{Preamble}
\include{Background}
\include{ANS}
\include{PLC}
\include{Vectorizing}
\include{HiLLoC}
\include{Conclusions}
\include{Appendices}

\printbibliography
\addcontentsline{toc}{chapter}{Bibliography}

\end{document}

%% file: MainPackages.tex
\usepackage{emptypage}
\usepackage{graphicx}
\usepackage{float}

\usepackage{setspace}
\usepackage{multirow}

\usepackage[format=hang,font=small,labelfont=bf]{caption}

\usepackage{j} %

\usepackage{listings}
\theoremstyle{definition}
\newtheorem{definition}{Definition}
\theoremstyle{plain}
\newtheorem{theorem}{Theorem}

\newcommand{\push}{\texttt{push}}
\newcommand{\pop}{\texttt{pop}}

\usepackage{subcaption}

\usepackage{tablefootnote}

%% file: Figures.tex
\usepackage{tikz}
\usetikzlibrary{calc,math,decorations.pathreplacing,arrows,positioning,shapes,
                quotes,arrows.meta}
\newlength{\barwidth}
\newcommand{\drawaxes}[1]{
  \draw[gray]  (-.2, 0) -> (3.75 * \barwidth + .2cm, 0);
  \draw                    (3.75 * \barwidth + .2cm, 0) node[anchor=west] {$x$};
  \draw[gray]  (-.2, 0) -> (-.2, 5.2);
  \draw                    (-.2, 5.2) node[anchor=south] {$#1$};
  \foreach \x in {0.5, 1.5, 2.5, 3.5}{
    \draw[gray] (\x * \barwidth, -2pt) -- (\x * \barwidth, 0pt);
  }
  \foreach \x/\char in {0.5/a, 1.5/b, 2.5/c, 3.5/d}{
    \draw (\x * \barwidth, 0) node[anchor=north, text height=8] {\texttt\char};
  }
}

\newcommand{\ytick}[2]{%
  \draw[gray] (-.2cm - 2pt, #1) node[anchor=east, color=black] {#2} -- (-.2cm + 0pt, #1);
}

\newcommand{\drawbar}[3]{
  \draw[#3] (#1, 0) -- (#1, 12.5cm * #2);
}

\newcommand{\drawpmf}{
\begin{tikzpicture}
  \drawaxes{P(x)}
  \foreach \y in {0.125, 0.25, 0.375}{\ytick{12.5 * \y}{$\y$}}
  \drawbar{0.5\barwidth}{0.125}{}
  \drawbar{1.5\barwidth}{0.25}{}
  \drawbar{2.5\barwidth}{0.375}{}
  \drawbar{3.5\barwidth}{0.25}{}
\end{tikzpicture}
}

\newcommand{\drawapprox}[2]{
\begin{tikzpicture}
  \tikzmath{
    \splus = #1 + 1;
    \yunit = 12.5 / \splus;
  }
    \drawaxes{s'}
  \foreach \y [parse=true, count=\s] in {\yunit, 2 * \yunit, ..., 5.2}{
    \ytick{\y}{}
    \draw (-.2cm, \y - \yunit / 2) node[anchor=east]
      {\pgfmathparse{int(\s - 1)}#2$\pgfmathresult$};
  }
  \foreach \x/\freq/\start in {0/1/0, 1/2/1, 2/3/3, 3/2/6}{
    \foreach \y [parse=true, count=\s] in {\yunit, 2 * \yunit, ..., 5.2}{
      \tikzmath{
        \snew = int(8 * int((\s - 1) / \freq) + mod((\s - 1), \freq) + \start);
        if \snew<=#1 then {\toprint = \snew;} else {let \toprint = ;};
        }
      \draw ($(0, -\yunit / 2) + 0.5*(\barwidth,0)
      + \s*(0, \yunit) + \x*(\barwidth, 0)$) node {#2$\toprint$};
    }
  }
\end{tikzpicture}
}

\newlength{\intervalwidth}
\newlength{\gap}
\newlength{\lineheight}
\newcommand{\drawinterval}{
\begin{tikzpicture}
  \draw (-30pt, 0) node [anchor=east] {$s$};
  \draw (0, -.5\lineheight) -- (0, .5\lineheight);
  \draw (\intervalwidth, -.5\lineheight) -- (\intervalwidth, .5\lineheight);
  \foreach \s in {1, ..., 7}{
    \draw (\s * 0.125 * \intervalwidth, -.5\lineheight)
      -- (\s * 0.125 * \intervalwidth, .5\lineheight);
  }
  \foreach \y in {0, ..., 7}{
    \tikzmath{
      \x = (\y * 0.125 + 0.0625);
      \r = int(\y + 64);
    }
    \draw (\x\intervalwidth, 0) node {$\r$};
  }

  \draw (-30pt, -\lineheight -\gap) node [anchor=east] {$s \bmod 2^{r}$};
  \draw (0, -.5\lineheight - \gap) -- (0, -1.5\lineheight -\gap);
  \draw (\intervalwidth, -.5\lineheight - \gap)
    -- (\intervalwidth, -1.5\lineheight - \gap);
  \foreach \s in {1, ..., 7}{
    \draw (\s * 0.125 * \intervalwidth, -.5\lineheight - \gap)
      -- (\s * 0.125 * \intervalwidth, -1.5\lineheight - \gap);
  }
  \foreach \s in {0, ..., 7}{
    \tikzmath{
      \x = (\s * 0.125 + 0.0625);
    }
    \draw (\x\intervalwidth, -\lineheight - \gap) node  {$\s$};
  }

  \draw (-30pt, -2\gap - 2\lineheight) node [anchor=east] {$x$};
  \draw (0, -2\gap - 1.5\lineheight) -- (0, -2\gap - 2.5\lineheight - 6pt);
  \draw (\intervalwidth, -2\gap - 1.5\lineheight)
    -- (\intervalwidth, -2\gap - 2.5\lineheight - 6pt);
  \draw (0, -2\gap - 2.5\lineheight - 6pt) node [anchor=north] {$0$};
  \draw (\intervalwidth, -2\gap - 2.5\lineheight - 6pt) node [anchor=north] {$1$};
  \draw (0.125  \intervalwidth, -2\gap - 1.5\lineheight)
    -- (0.125 * \intervalwidth, -2\gap - 2.5\lineheight);
  \draw (0.375  \intervalwidth, -2\gap - 1.5\lineheight)
    -- (0.375 * \intervalwidth, -2\gap - 2.5\lineheight);
  \draw (0.75  \intervalwidth, -2\gap - 1.5\lineheight)
    -- (0.75 * \intervalwidth, -2\gap - 2.5\lineheight);
  \draw (0.06125\intervalwidth, -2\gap - 2\lineheight)
    node [text height=1ex] {\texttt{a}};
  \draw (0.25   \intervalwidth, -2\gap - 2\lineheight)
    node [text height=1ex] {\texttt{b}};
  \draw (0.5625 \intervalwidth, -2\gap - 2\lineheight)
    node [text height=1ex] {\texttt{c}};
  \draw (0.875  \intervalwidth, -2\gap - 2\lineheight)
    node [text height=1ex] {\texttt{d}};

\end{tikzpicture}
}

\newlength{\tbase}
\newcommand{\drawmessage}{
\begin{tikzpicture}
  \node at (0, 0) (s) {\texttt{0110001001110010}};
  \node at (0, -0.85) {\large\(s\)};
  \draw [{|-|}] (-1.6, 0.5) -- (1.6, 0.5);
  \node at (0, 0.8) {\(r_s\)};

  \node at (4.5, \tbase) {\texttt{01100010}};
  \node at (4.5, \tbase + 11) {\texttt{10100011}};
  \node at (4.5, \tbase + 30) {\(\vdots\)};
  \node at (4.5, \tbase + 43) {\texttt{11011001}};
  \draw [{|-|}] (4.5 - 0.8, \tbase + 58) -- (4.5 + 0.8, \tbase + 58);
  \node at (4.5, \tbase + 66) {\(r_t\)};
  \node at (4.5, -0.85) {\large\(t\)};
  \draw [{|-|}] (4.5 + 1.2, \tbase - 4) -- (4.5 + 1.2, \tbase + 43 + 4);
  \node at (4.5 + 1.6, \tbase + 21.5) {\(\abs{t}\)};
\end{tikzpicture}
}

\newcommand{\drawvectormessage}{
\begin{tikzpicture}
  \node at (0, -0.85) {\large\(s\)};
  \node at (0, \tbase + 53) {\(r_s\)};
  \draw [{|-|}] (-1.57, \tbase + 45) -- (1.57, \tbase + 45);
  \node at (0, \tbase + 30) {\texttt{0010110011010011}};
  \node at (0, \tbase + 19) {\(\vdots\)};
  \node at (0, \tbase)      {\texttt{0110001001110010}};
  \draw [{|-|}] (1.9, \tbase - 4) -- (1.9, \tbase + 30 + 4);
  \node at (2.3, \tbase + 15) {\(K\)};

  \node at (4.5, \tbase) {\texttt{01100010}};
  \node at (4.5, \tbase + 11) {\texttt{10100011}};
  \node at (4.5, \tbase + 30) {\(\vdots\)};
  \node at (4.5, \tbase + 43) {\texttt{11011001}};
  \draw [{|-|}] (4.5 - 0.77, \tbase + 58) -- (4.5 + 0.77, \tbase + 58);
  \node at (4.5, \tbase + 66) {\(r_t\)};
  \node at (4.5, -0.85) {\large\(t\)};
  \draw [{|-|}] (4.5 + 1.2, \tbase - 4) -- (4.5 + 1.2, \tbase + 43 + 4);
  \node at (4.5 + 1.6, \tbase + 21.5) {\(\abs{t}\)};
\end{tikzpicture}
}

\tikzset{
  scuboid/.pic={
    \tikzset{%
      every edge quotes/.append style={midway, auto},
      /scuboid/.cd,
      #1
    }
    \draw [densely dashed, pic actions]
    (0,0,0) -- ++(-\cubescale*\cubex,0,0)
    -- ++(0,0,-\cubescale*\cubez)
    -- ++(\cubescale*\cubex,0,0)
    -- ++(0,0,\cubescale*\cubez)
    -- cycle;
    \draw (0,0,0) -- ++(0,-\cubescale*\cubey,0)
    -- ++(-\cubescale*\cubex,0,0)
    -- ++(0,\cubescale*\cubey,0);

    \draw (0,-\cubescale*\cubey,0)
    -- ++(0,0,-\cubescale*\cubez)
    -- ++(0,\cubescale*\cubey,0);
  },
  /scuboid/.search also={/tikz},
  /scuboid/.cd,
  width/.store in=\cubex,
  height/.store in=\cubey,
  depth/.store in=\cubez,
  units/.store in=\cubeunits,
  scale/.store in=\cubescale,
  width=10,
  height=10,
  depth=10,
  units=cm,
  scale=.1,
}

\newcommand{\drawcraystackdiagram}{
\begin{tikzpicture}
    \pic at (0,0) {scuboid={width=0, height=0, depth=0}};

    \node at (2.85, -1.15) {$x \rightarrow p(\cdot \given z)$};

    \node at (8.5, -1.15) {$z \rightarrow p(\cdot)$};

    \pic [fill=gray!20] at (5.1, -3) {scuboid={width=33, height=5, depth=32}};
    \pic [fill=gray!20] at (11.5, -3) {scuboid={width=50, height=5, depth=32}};

     \draw [-{Latex[length=1.5mm,width=2mm]}] (4,0) -- (4,-2.5);
    \draw [-{Latex[length=1.5mm,width=2mm]}] (9.5,0) -- (9.5,-2.5);

    \node[inner sep=0pt] at (4.05,0)
    {\includegraphics[width=.325\textwidth]{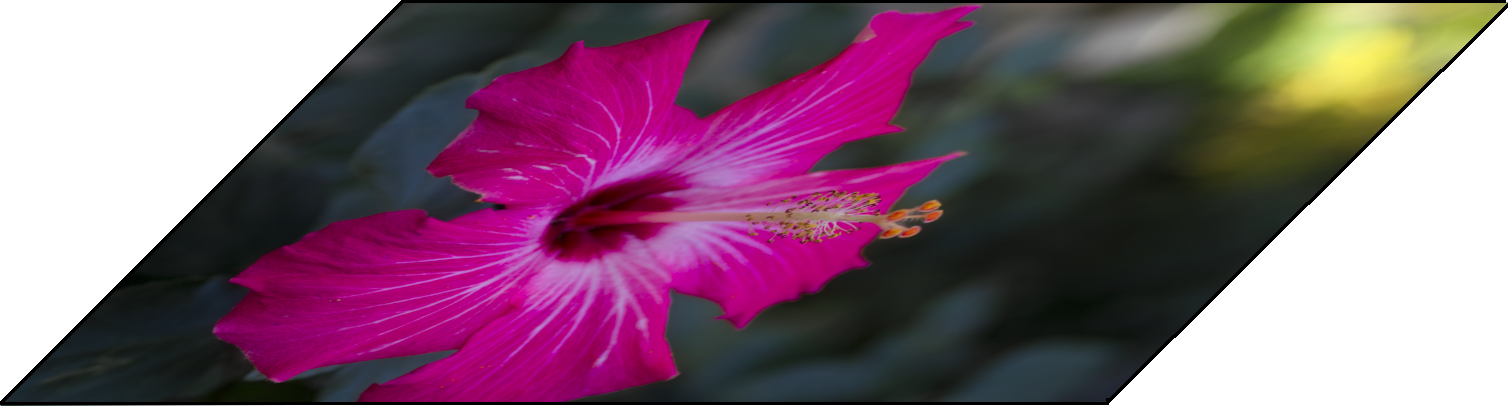}};

    \node[inner sep=0pt] at (9.575,0)
    {\includegraphics[width=.45\textwidth]{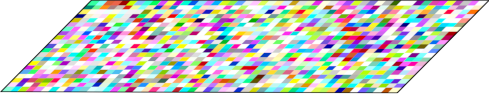}};

    \draw [decorate,decoration={brace,amplitude=10pt,mirror}] (1.25,-1.8) -- (1.25,-4.5);
    \node[text width=1.5cm,align=center] at (0,-3.15) {Vectorized ANS coder};

    \node (xbelow) at (4,-3.375) {};
    \node (zbelow) at (9.5,-3.375) {};
    \node (ztemp) at (8,-4.0) {};
    \node (ztemp2) at (8.25,-4.0) {};
    \node (ztemp3) at (6.65,-4.) {};
    \node (tailleft) at (5.8,-4.5) {};
    \node (tailleft2) at (5.8,-4.45) {};

    \draw [-{Latex[length=1.5mm,width=2mm]}] (xbelow) to [out=-90,in=180] (tailleft2);
    \draw (zbelow) to [out=-90,in=0] (ztemp);
    \draw (ztemp2) -- (6.5, -4);
    \draw [-{Latex[length=1.5mm,width=2mm]}] (ztemp3) to [out=180, in=135] (tailleft);

    \draw[fill=black!75] (5.75,-4.5) rectangle ++(0.25,0.25);
    \draw[fill=black!75] (6,-4.5) rectangle ++(0.25,0.25);
    \draw[] (6.25,-4.5) rectangle ++(0.25,0.25);
    \draw[fill=black!75] (6.5,-4.5) rectangle ++(0.25,0.25);
    \draw[] (6.75,-4.5) rectangle ++(0.25,0.25);
    \draw[] (7,-4.5) rectangle ++(0.25,0.25);
    \draw[] (7.25,-4.5) rectangle ++(0.25,0.25);
    \draw[] (7.5,-4.5) rectangle ++(0.25,0.25);
    \draw[fill=black!75] (7.75,-4.5) rectangle ++(0.25,0.25);
    \draw[fill=black!75] (8,-4.5) rectangle ++(0.25,0.25);
    \draw[] (8.25,-4.5) rectangle ++(0.25,0.25);
    \draw[fill=black!75] (8.5,-4.5) rectangle ++(0.25,0.25);
    \draw[] (8.75,-4.5) rectangle ++(0.25,0.25);
    \draw[fill=black!75] (9,-4.5) rectangle ++(0.25,0.25);
    \draw[fill=black!75] (9.25,-4.5) rectangle ++(0.25,0.25);
    \draw[] (9.5,-4.5) rectangle ++(0.25,0.25);

    \draw (9.75,-4.5) -- (10, -4.5);
    \draw (9.75,-4.25) -- (10, -4.25);

    \node at (10.25, -4.375) {...};

\end{tikzpicture}
}

\newcommand{\drawhillocgenerative}{
\begin{tikzpicture}
\tikzstyle{var}=[circle,draw=black,minimum size=7mm]
\tikzstyle{latent3}  =[]
\tikzstyle{latent2}  =[]
\tikzstyle{latent1}  =[]
\tikzstyle{observed}=[fill=gray!50]
    \node[var,latent3]   (l3)               {$z_3$};
    \node[var,latent2]   (l2) [below of=l3] {$z_2$};
    \node[var,latent1]   (l1) [below of=l2] {$z_1$};
    \node[var,observed] (x) [below of=l1]   {$x$};
    \draw [->] (l3) -- (l2);
    \draw [->] (l2) -- (l1);
    \draw [->] (l3) to[out=-135,in=135] (l1);
    \draw [->] (l1) -- (x);
    \draw [->] (l2) to[out=-45,in=45] (x);
    \draw [->] (l3) to[out=-135,in=135] (x);
\end{tikzpicture}
}

\newcommand{\drawhillocinference}{
\begin{tikzpicture}
\tikzstyle{var}=[circle,draw=black,minimum size=7mm]
\tikzstyle{latent3}  =[]
\tikzstyle{latent2}  =[]
\tikzstyle{latent1}  =[]
\tikzstyle{observed}=[fill=gray!50]
    \node[var,latent3]   (l3)               {$z_3$};
    \node[var,latent2]   (l2) [below of=l3] {$z_2$};
    \node[var,latent1]   (l1) [below of=l2] {$z_1$};
    \node[var,observed] (x) [below of=l1]   {$x$};
    \draw [->] (l3) -- (l2);
    \draw [->] (l2) -- (l1);
    \draw [->] (l3) to[out=-135,in=135] (l1);
    \draw [dashed,->] (x) -- (l1);
    \draw [dashed,->] (x) to[out=45,in=-45] (l2);
    \draw [dashed,->] (x) to[out=45,in=-45] (l3);
\end{tikzpicture}
}

\newcommand{\drawbitswapgenerative}{
\begin{tikzpicture}
\tikzstyle{var}=[circle,draw=black,minimum size=7mm]
\tikzstyle{latent3}  =[]
\tikzstyle{latent2}  =[]
\tikzstyle{latent1}  =[]
\tikzstyle{observed}=[fill=gray!50]
    \node[var,latent3]   (l3)               {$z_3$};
    \node[var,latent2]   (l2) [below of=l3] {$z_2$};
    \node[var,latent1]   (l1) [below of=l2] {$z_1$};
    \node[var,observed] (x) [below of=l1]   {$x$};
    \draw [->] (l3) -- (l2);
    \draw [->] (l2) -- (l1);
    \draw [->] (l1) -- (x);
\end{tikzpicture}
}

\newcommand{\drawbitswapinference}{
\begin{tikzpicture}
\tikzstyle{var}=[circle,draw=black,minimum size=7mm]
\tikzstyle{latent3}  =[]
\tikzstyle{latent2}  =[]
\tikzstyle{latent1}  =[]
\tikzstyle{observed}=[fill=gray!50]
    \node[var,latent3]   (l3)               {$z_3$};
    \node[var,latent2]   (l2) [below of=l3] {$z_2$};
    \node[var,latent1]   (l1) [below of=l2] {$z_1$};
    \node[var,observed] (x) [below of=l1]   {$x$};
    \draw [->] (l2) -- (l3);
    \draw [->] (l1) -- (l2);
    \draw [dashed,->] (x) -- (l1);
\end{tikzpicture}
}

%% file: Preamble.tex
\title{Lossless Compression with Latent Variable Models}
\author{James Townsend}
\department{Department of Computer Science}

\maketitle
\makedeclaration

\begin{abstract} %
  We develop a simple and elegant method for lossless compression using latent
  variable models, which we call ‘bits back with asymmetric numeral systems’
  (BB-ANS)\@. The method involves interleaving encode and decode steps, and
  achieves an optimal rate when compressing batches of data. We demonstrate it
  firstly on the MNIST test set, showing that state-of-the-art lossless
  compression is possible using a small variational autoencoder (VAE) model. We
  then make use of a novel empirical insight, that fully convolutional
  generative models, trained on small images, are able to generalize to images
  of arbitrary size, and extend BB-ANS to hierarchical latent variable models,
  enabling state-of-the-art lossless compression of full-size colour images
  from the ImageNet dataset. We describe ‘Craystack’, a modular software
  framework which we have developed for rapid prototyping of compression using
  deep generative models.
\end{abstract}

\begin{impactstatement}
The research on which this thesis is based has been published at two top tier
machine learning conferences \citep{townsend2019,townsend2020}. It has also
been included in the prestigious and highly popular `Deep Unsupervised
Learning' course, taught at UC Berkeley, taking up roughly one hour of lecture
material.  The lecture includes a number of verbatim extracts and diagrams from
the paper \citet{townsend2019}, upon which \Cref{chap:plc} is based. At the
time of writing, a video of the lecture had been viewed over 5 thousand times
on YouTube\footnote{See \jurl{youtu.be/0IoLKnAg6-s?t=5681}.}. A number of
recent publications by other researchers have been based upon our work,
including \citet{hoogeboom2019,ho2019,kingma2019,vandenberg2020}.

There is great potential for translation of this work into improvements to
widely used production compression systems, both in general purpose codecs
which are used by billions of people on a daily basis as they browse the
internet, and more bespoke systems for specific applications such as
compression of medical or astronomical images. We have sought to maximise
long-term impact by publishing the tutorial paper \citet{townsend2020a}, which
is intended as an accessible introduction to the ideas in this thesis; as well
as releasing open source software for easy prototyping of future ideas in the
area which this thesis covers, namely lossless compression using deep
generative models\footnote{Software available at
\jurl{github.com/j-towns/craystack}, mirrored at
\jurl{doi.org/10.5281/zenodo.4707276}.}.
\end{impactstatement}

\begin{acknowledgements}
  Thanks to Mark Herbster for encouraging me to start a PhD, without Mark I
  certainly would not have started on this journey, and I can't thank him
  enough for gently nudging me in this direction. Thanks to my co-authors Tom
  Bird and Julius Kunze,  they are both gifted researchers and working with
  them has been an absolute joy. Thanks to my supervisor David Barber and to
  others who have given advice and feedback on this work, particularly Paul
  Rubenstein and Raza Habib. Thanks to the examiners, Matt Kusner and Iain
  Murray, for an exciting viva and excellent, thorough feedback. A million
  thanks to my family, to my partner Maud and to my best friend Anthony, for
  all of their love and support.
\end{acknowledgements}

\begin{preface}
The connections between information theory and machine learning have long been
known to be deep. The two fields are so closely related that they have been
described as `two sides of the same coin' by \citet{mackay2003}, who insists,
in the preface to his book, that they `belong together'. One particularly
elegant connection is the essential equivalence between probabilistic models of
data and lossless compression methods. The source coding theorem
\citep{shannon1948} can be thought of as the fundamental theorem describing
this idea, and Huffman coding \citep{huffman1952}, arithmetic coding
\citep[AC;][]{witten1987} and the more recently developed asymmetric numeral
systems \citep[ANS;][]{duda2009} are actual algorithms for implementing
lossless compression, given some kind of probabilistic model.

The field of machine learning has experienced an explosion of activity in recent
years, and, amongst other things, we have seen major breakthroughs in
probabilistic modelling of high dimensional data. Recurrent neural networks and
autoregressive models based on masked convolution have been shown to be
effective generative models for images, audio and natural language
\citep{graves2014, vandenoord2016a, vandenoord2016}. These models are slow to
sample from, at least in a na\"ive implementation, which means that
decompression using these models is similarly slow\footnote{
  Sampling in autoregressive models based on masked convolution can be sped up
  drastically using dynamic programming \citep{lepaine2016, ramachandran2017}.
  However, there is not yet a general implementation of this method and at
  present a lot of developer time is required to hand-implement fast sampling
  for each individual model.
}, however they do tend to achieve state
of the art test log-likelihoods, and hence state of the art compression rates.

Another significant, recently invented type of probabilistic generative model
is the variational autoencoder (VAE), first presented in \citet{kingma2014} and
\citet{rezende2014}. VAEs are latent variable models which use a neural network
for efficient posterior inference and where the generative model and posterior
inference network are jointly trained, using stochastic gradient descent on the
variational free energy, an objective function which we refer to in this work
as the `evidence lower bound' (ELBO). VAE models have been shown to obtain
competitive (though usually not state of the art) log-likelihoods, and sampling
from them tends to be much faster than it is from autoregressive and recurrent
models.

In the last five years we have seen a number of papers covering applications of
modern deep learning methods to \emph{lossy} compression. \citet{gregor2015}
discusses applications of a VAE to compression, with an emphasis on lossy
compression. \citet{balle2017, theis2017, balle2018, minnen2018} all implement
lossy compression using (variational) auto-encoder style models, and
\citet{tschannen2018} train a model for lossy compression using a GAN-like
objective. Applications to \emph{lossless} compression were less well covered
in works published prior to \citet{townsend2019}, upon which \Cref{chap:plc} is
based.

The classic lossless compression algorithms mentioned above (Huffman coding,
AC, and ANS) do not naturally cater for latent variable models. However there
is a method, known as `bits back coding', which can be used to extend those
algorithms to cope with such models \citep{wallace1990, hinton1993}. Bits back
coding was originally introduced as a purely theoretical construct, but was
later implemented, in a primitive form, by \citet{frey1996}.

There is a fundamental incompatibility between the bits back method and the AC
scheme upon which the primitive implementation was based. A workaround is
suggested by \citet{frey1997}, but this leads to a sub-optimal compression rate
and an overly complex implementation.  The central theoretical contribution of
this thesis is a simple and elegant solution to the issue just mentioned, which
involves implementing bits back using ANS instead of AC. We term the new coding
scheme `Bits Back with ANS' (BB-ANS).

Our scheme improves on early implementations of bits back coding in terms of
compression rate and code complexity, allowing for efficient lossless
compression of batches of data with deep latent variable models.  We are also
the first to implement a method, first suggested by \citet{mackay2003} for
using bits back coding with continuous latent variables. In \Cref{chap:plc} we
demonstrate the efficiency of BB-ANS by losslessly compressing the MNIST
dataset with a VAE.\looseness=-1

We find that BB-ANS with a VAE outperforms generic compression algorithms for
both binarized and raw MNIST, with a very simple model architecture. In
\Cref{chap:vectorizing} we lay the ground work for scaling BB-ANS up to larger
models, describing a method for vectorizing the underlying ANS coder, and
discussing the generic software which we have written, enabling other machine
learning researchers to easily prototype compression systems. In
\Cref{chap:hilloc} we present an extension of BB-ANS to hierarchical latent
variable models. We show that it is possible to use a fully convolutional model
to compress images of arbitrary shape and size, and use this technique with
BB-ANS to achieve a 6.5\% improvement over the previous state of the art
compression rate on full-size images from the ImageNet dataset.

In our experiments, we benchmark BB-ANS on image data because this is a
well-studied domain for deep generative models. The (non-learned) image codecs
which we benchmark against are Portable Network Graphics (PNG), WebP lossless,
and Free Lossless Image Format\footnote{FLIF is being subsumed by the JPEG XL
standard which is currently under development. JPEG XL is more general and
achieves better lossless compression rates than FLIF, see
\cite{alakuijala2019}.} (FLIF), which were first released in 1997, 2012, and
2015, respectively. All three of these codecs achieve compression using local,
low-dimensional prediction. They adapt (effectively `learning') as more pixels
within a single image are processed. The broad approach taken in this thesis,
which can be applied to data types other than images, and with models other
than VAEs, is to spend a large amount of computation time adapting (training) a
model on a generic dataset such as ImageNet, \emph{before} it is presented with
an image which it is tasked with compressing.

I would argue that the early results from using this approach, which are
presented in this thesis, as well as the other recent works, are promising. In
the period since the publication of \citet{townsend2019}, there have been a
number of articles on this topic, most of which make use of ideas from our
work. \citet{mentzer2019} was published just after \citet{townsend2019}, and
demonstrates learned lossless compression with a relatively weak model,
achieving excellent run-times but failing to outperform existing methods in
terms of compression rate. \citet{kingma2019} builds directly on our work,
proposing an extension of BB-ANS to hierarchical models; we compare this to our
own extension in \Cref{chap:hilloc}. Very recent work by \citet{ruan2021} also
directly extends the methods in this thesis, using Monte Carlo methods to
improve the compression rate.  \citet{hoogeboom2019,ho2019,vandenberg2020} all
use ANS with flow models to do lossless compression. They show excellent
lossless compression performance on \(32\times32\) and \(64\times64\) images
but do not manage to scale their methods up to large images as we have.  It is
still an open question whether this potential paradigm shift can be exploited
in the production-grade systems which are used by billions of people every day,
and whether the gains will be as impressive in other domains, such as audio and
video.

\section*{Structure of the thesis}
This thesis is comprised of a background chapter (\Cref{chap:bg}), followed by
four `core' chapters (\Cref{chap:ans,chap:plc,chap:vectorizing,chap:hilloc}).
It is designed to be read by someone who has a little background knowledge in
information theory and coding theory (\cite{mackay2003}, Chapters 4-6, provides
an ideal introduction to the necessary topics), and who is already familiar
with the types of modern generative models discussed above, particularly
variational autoencoders (VAEs). A brief outline of the core chapters:
\begin{enumerate}
\setcounter{enumi}{1}
\item \emph{\nameref{chap:ans}}

  We introduce ANS coding, proving worst-case bounds on its performance. We
  include diagrams and pseudocode to assist the reader in their understanding.
  We also provide a 50 line working Python implementation to accompany this
  chapter, which can be found in \Cref{app:ans-example}.  This chapter is based
  on our tutorial paper, \citet{townsend2020a}. The ANS algorithm was first
  presented in \citet{duda2009}.

\item \emph{\nameref{chap:plc}}

  We present our novel approach to bits-back coding and demonstrate its
  performance by compressing the MNIST test set with a small VAE model. We call
  this method `bits-back with asymmetric numeral systems' (BB-ANS). We discuss
  a number of potential issues with the method and also discuss ways to improve
  this simple prototype system. This chapter is based on our paper,
  \citet{townsend2019}.

\item \emph{\nameref{chap:vectorizing}}

  We describe a method for implementing a vectorized ANS coder, drawing on
  earlier work by \citet{giesen2014}. We describe `Craystack', a software tool
  which we have developed which aims to provide a flexible, user-friendly API
  for machine learning practitioners who wish to prototype compression systems.
  We discuss future directions for Craystack.

\item \emph{\nameref{chap:hilloc}}

  We demonstrate that the BB-ANS method can be scaled up to hierarchical VAEs
  and large, colour images from the ImageNet test set. The method achieves
  state of the art compression on a randomly selected subset of 2,000 images
  from the ImageNet dataset. We discuss the challenges and solutions which were
  necessary to achieve this scale-up. This chapter is based on our paper,
  \citet{townsend2020}.
\end{enumerate}
Although each chapter in some ways builds on all of the previous, it should be
possible to read and understand (at least on a high level)
\Cref{chap:vectorizing} without reading \Cref{chap:plc}, and \Cref{chap:hilloc}
without reading \Cref{chap:vectorizing}.  \Cref{fig:chapgraph} visualizes this
approximate dependency structure between the core chapters.

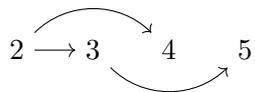
\begin{figure}[ht]
\centering
\begin{tikzpicture}
  \node (two)                    {2};
  \node (three) [right of=two]   {3};
  \node (four)  [right of=three] {4};
  \node (five)  [right of=four]  {5};

  \draw [->] (two)   to (three);
  \draw [->] (two)   to [bend left=45] (four);
  \draw [->] (three) to [bend right=45] (five);
\end{tikzpicture}
\caption{
  Dependencies between the core chapters of this thesis. We recommend not to
read a chapter before reading its parents in this graph.}\label{fig:chapgraph}
\end{figure}
\end{preface}

\setcounter{tocdepth}{2}
\tableofcontents
\listoffigures
\listoftables

%% file: Background.tex
\chapter{Background}\label{chap:bg}
This chapter aims to give a concise overview of the necessary background
material for understanding the rest of the thesis and its context within the
existing compression literature. We begin, in \Cref{sec:pgms}, by describing
probabilistic generative models, and specifically latent variable models, which
are a central tool in the compression methods developed in later chapters.
Then in \Cref{sec:sct} we review some theoretical background material in
compression, and in particular we define lossless compression and give a
statement of the source coding theorem \citep{shannon1948}. In
\Cref{sec:old-coding-techniques}, we outline a few of the most commonly used,
generic, lossless compression methods. Finally, in
\Cref{sec:practical-image-compression}, we motivate lossless compression of
images and give an overview of two existing codecs, comparing their approach to
ours.

\section{Probabilistic generative models}\label{sec:pgms}
A probabilistic model describes a variable (or set of variables) whose value is
random. Such a model may be characterised by a probability `mass function' \(P
\colon \mathcal{X} \rightarrow \mathbb{R}\), where \(\mathcal{X}\) is a
countable (usually finite) set. The function \(P\) must satisfy two properties
\begin{enumerate}
  \item \(\forall x \in \mathcal{X},\quad P(x) \geq 0\)
  \item \(\sum_{x\in\mathcal{X}} P(x) = 1\).
\end{enumerate}

We say that the distribution defined by \(P\) is `discrete'. This definition
can be generalized to sets \(\mathcal{X}\) which are Lebesgue measurable (such
as the set \(\mathbb{R}\) of real numbers). Then instead of a mass function the
distribution is characterised by a `density function' \(p\colon\mathcal{X}
\rightarrow \mathbb{R}\), with
\begin{enumerate}
  \item \(\forall x \in \mathcal{X},\quad p(x) \geq 0\)
  \item \(\int_{x\in\mathcal{X}} p(x) = 1\)
\end{enumerate}
and we say that the distribution is `continuous', rather than discrete. As we
will see in \Cref{chap:ans}, we can efficiently compress outcomes from a
discrete distribution if they can be broken down into a sequence of `symbols',
i.e.\ if we can write \(x=x_1,x_2,...\) with \(P(x) = \prod_t P(x_t\given
x_1,\ldots,x_{t-1})\) and if we are able to compute the cumulative distribution
function (CDF) and its inverse for each of the factors \(P(x_t\given
x_1,\ldots,x_{t-1})\).

In practice, a mass function or density function will usually have one or more
tunable parameters. We usually use \(\theta\) to denote the vector containing a
model's parameters and write \(P(x;\theta)\) for the parametrized mass
function. In the context of modern machine learning, the parameters of a
probabilistic generative model are usually tuned (or `trained'), by optimizing
the `log-likelihood' function over a set of example data, referred to as the
`training set'. That is, by solving the following optimization problem:
\begin{equation}\label{eq:ml-objective}
  \hat\theta = \argmax_\theta \sum_{x\in{X}} \log P(x; \theta),
\end{equation}
where \(X\) is a set of example data. This technique is known as `maximum
likelihood' (ML) learning. If \(P\) can be tractably computed, then its
gradient with respect to \(\theta\) can usually be computed using automatic
differentiation (AD), and the log-likelihood optimized using stochastic
gradient descent (SGD). After optimization the model is usually evaluated on a
held-out set of examples called the `test set'.

In the last five years, maximum likelihood training has been scaled to models
with billions of parameters, and data sets with millions of examples, utilizing
parallel graphics processing unit (GPU) based hardware for efficient training.
Two particularly famous examples of systems which use this technique are WaveNet
\citep{vandenoord2016}, which is a probabilistic generative model for audio,
used in Google's speech synthesis applications, and GPT-3 \citep{brown2020}, a
natural language model with 175 billion parameters.  Models for images which
are trained using maximum likelihood are also fast approaching photo-realism in
the samples which they generate. Recent examples include
\citet{menick2018,jun2020,child2020}.

\subsection{Latent variable models}\label{sec:lvms}
Latent variable models are a class of probabilistic generative model which
involve unobserved, or `latent', variables. Their mass function is defined
implicitly by an integral, or, to use the terminology of probability theory, by
`marginalizing' a latent variable:
\begin{equation}\label{eq:lvm}
  P(x) := \int_z P(x\given z) p(z) dz
\end{equation}
The distribution specified by \(p(z)\) is referred to as the `prior', and the
forward probability \(P(x\given z)\) the `likelihood' (note likelihood here has
a slightly different meaning to the definition in the previous section). We
usually choose prior and likelihood distributions which are straightforward to
sample from, and thus exact sampling from \(P(x)\) is straightforward by first
sampling \(z\) from the prior and then sampling \(x\) from the likelihood,
conditioned on the sampled \(z\).

A popular class of latent variable models, which we use to demonstrate our
methods in \Cref{chap:plc,chap:hilloc}, are called `variational auto-encoders'
(VAEs), first introduced in \citet{kingma2014,rezende2014}. They usually use a
simple, fixed prior distribution, such as a multivariate Gaussian with mean
zero and with covariance equal to the identity matrix. The likelihood function
\(P(x\given z)\) is usually of the form \(P_\mathrm{simple}(x\given f(z;
\theta))\), where \(f\) is a multi-layer neural network (i.e.\ a composition of
differentiable, parametrized functions), and \(P_\mathrm{simple}\) is a mass
function which is straightforward to compute, and has the appropriate support.
For discrete data, a discretized logistic distribution is often used (we give
details of the distributions we used in our experiments in later chapters).

For VAE models the integral \cref{eq:lvm} cannot easily be computed, and thus
direct maximum likelihood training is not possible. To train a VAE, we instead
optimize a variational lower bound on \(\log P(x; \theta)\), called the
`evidence lower-bound' (ELBO), sometimes referred to as the `variational free
energy'. It is defined as
\begin{equation}\label{eq:bg-elbo-def}
  \mathcal{L}(x; \theta, \phi) = \int_z q(z\given x;\phi) \log \frac{P(x\given z;
    \theta)p(z)}{q(z\given x; \phi)} dz.
\end{equation}
We refer to the newly introduced density function \(q\) as the variational
posterior, or the approximate posterior, and the new parameters \(\phi\) as the
variational parameters. The fact that \(\mathcal{L}(x; \theta, \phi) \leq P(x;
\theta)\) follows directly from Jensen's inequality.

The approximate posterior \(q\) can, in principal, be any distribution, but in
VAEs it is common to use a parametrization similar to that of the likelihood,
i.e.\ a function with the form \(q_\mathrm{simple}(z\given g(x; \phi))\), where
\(g\) is a multi-layer neural network and \(q_\mathrm{simple}\) is usually a
Gaussian distribution with a diagonal covariance matrix. Exact sampling from
\(q\) is then straightforward, and this allows us to compute an unbiased
Monte Carlo estimate
of \(\mathcal{L}\):
\begin{equation}
  \hat{\mathcal{L}}(x; \theta, \phi) = \log \frac{P(x\given z;
  \theta)p(z)}{q(z\given x; \phi)} \text{ where } z\leftarrow q(\cdot\given x;
  \phi).
\end{equation}
The shorthand notation \(z\leftarrow q(\cdot\given x; \phi)\) means \(z\) is
sampled from the approximate posterior distribution \(q(z\given x; \phi)\).

If the process used to generate \(z\) from \(q\) is differentiable with respect
to \(\phi\), then the function \(\hat{\mathcal{L}}\) can be differentiated with
respect to \(\theta\) and \(\phi\) (we usually use automatic differentiation
tools to do this), and SGD can be used to optimize
\(\mathcal{L}\).\looseness=-1

Variational auto-encoders are reasonably straightforward to train and sample
from, and since 2014 there have been a huge number of papers presenting
different versions, with a general trend towards models with more parameters,
better samples, and more accurate density estimation.  Rather than survey this
extensive literature we simply point to the two most recent examples of works
which have pushed this envelope, \citet{maaloe2019,vahdat2020}, both of which
demonstrate VAE image models with samples that, to the human eye, appear
similar to real examples.

For the algorithms introduced later, we will need to be able to compress
outcomes from the prior, likelihood and posterior of a latent variable model.
As mentioned in \Cref{sec:pgms}, this means their mass functions must be
factorizable into a product of conditionals in such a way that we are able to
compute the CDF and its inverse under each conditional. Since, in a VAE, 
elements of the vectors \(x\) and \(z\) are usually modelled as independent
under the three relevant distributions, we can simply use the natural
factorization which has one factor for each element in the vector; it is also
common practice to use distributions for which the CDF and inverse CDF for each
element can easily be computed. Lossless compression is only possible for
discrete (rather than continuous) data, which would seem to render it
incompatible with the continuous latents typically used for VAEs.  However, it
turns out that it is straightforward to overcome this issue by quantizing
continuous latents, in a way which has only a very small effect on compression
rates. We will give more detail in \Cref{chap:plc} and \Cref{chap:hilloc}.

\subsection{Further background on probabilistic generative models}
For detailed background on probabilistic generative models,
\citet{mackay2003,bishop2006} are classic references. \citet{murphy2012} gives
a more modern, extremely thorough, overview of the field, and the even more
recent \citet{goodfellow2016} includes VAE models and detail on the various
neural network techniques which are useful for defining the functions
\(f(x;\theta)\) and \(g(z; \phi)\) mentioned above.

\section{Source coding}\label{sec:sct}
`Source coding', first formalized by \citet{shannon1948}, is the name we give
to the problem of trying to find lossless encodings for data which are drawn
from a random source. Source coding is more-or-less synonymous with `lossless
compression'. In this section we define basic terminology, then in
\Cref{sec:old-coding-techniques} we review some of the basic algorithms for
doing source coding. This is closely based on \citet{mackay2003} Chapters 4-7,
which we recommend as an introduction to these topics.

\subsection{Terminology and the source coding theorem}
We use the following definition for probability distributions, based on that
used by \citet{mackay2003}:
\begin{definition}\label{def:ensemble1}
  An \emph{ensemble} \(X\) is a triple \((x, \mathcal{A}_X, \mathcal{P}_X)\)
  where the \emph{outcome} \(x\) is the value of a random variable, taking on
  one of a set of possible values \(\mathcal{A}_X = \{a_1, \ldots, a_I\}\), and
  \(\mathcal{P}_X = \{p_1, \ldots, p_I\}\) are non-negative real-valued
  probability weights with each \(P(x=a_i) = p_i\) and therefore
  \(\sum_{i=1}^Ip_i = 1\).
\end{definition}
The following are the basic quantities of concern in source coding and
information theory in general. Here, and throughout the rest of the thesis, we
use `\(\log\)' for the base 2 logarithm, usually denoted `\(\log_2\)'.
\begin{definition}
  The \emph{Shannon information content} of an outcome \(x\) is
  \begin{equation}
    h(x=a_i) := \log \frac{1}{p_i}.
  \end{equation}
\end{definition}
\begin{definition}
  The \emph{entropy} of an ensemble \(X\) is
  \begin{equation}
    H(X) := \sum_i p_i\log\frac{1}{p_i}.
  \end{equation}
\end{definition}

The \emph{source coding theorem} demonstrates the relevance of the entropy as a
measure of a random variable's information content. We give an informal
statement of the theorem. For more detail, including a proof of the theorem,
see \citet{mackay2003}, Chapter 4.

\begin{theorem}[Source coding theorem, informal statement]\label{theorem:sct}
  \(N\) i.i.d.\ random variables each with entropy \(H(X)\) can be compressed
  into more than \(NH(X)\) bits with negligible risk of information loss, as
  \(N\rightarrow \infty\); but conversely, if they are compressed into fewer
  than \(NH(X)\) bits it is virtually certain that information will be lost.
\end{theorem}
This fundamental theorem specifies a theoretical limit of possibility for
lossless compression. Compression at rates close to the entropy is possible
(though may not be computationally feasible), and coding at better rates,
without loss of information, is not.

\subsection{Coding according to a model}
In the next section, some of the coding algorithms which we review assume
access to a probabilistic model, as defined in \Cref{sec:pgms}. In practical
settings, there is almost always some discrepency between the model used and the
true data generating distribution. To highlight this, we will use \(\tilde{p}_i
= P(x = a_i; \theta)\) and \(h(x=a_i;\theta) = \log 1/\tilde{p}_i\) for the
probabilities and information content \emph{according to the model}. The coding
algorithms are deterministic, and for a fixed input sequence their message
length has no dependence on the true generative distribution, but it will
depend on the model distribution, and in particular on \(h(x;\theta)\). In
particular, we will see that, for the three model-based algorithms that we
describe, we can write down a bound on compressed message length of the form
\begin{equation}\label{eq:true-length-simple-bound}
  l(x) \leq h(x;\theta) + \ldots
\end{equation}
This is useful, because the maximum likelihood objective, defined in
\cref{eq:ml-objective}, directly minimizes \(h(x;\theta)\) on training data.
Thus we might say that the log-likelihood is the correct objective function to
use to optimize a model for lossless compression (assuming any extra terms,
denoted `\(\ldots\)', are not too significant).

We can relate the above quantity to the entropy by taking the expectation of
\(h(x;\theta)\) over the true generating distribution
\begin{align}
  \E(h(x; \theta))
    &= \sum_i p_i \log\frac{1}{\tilde{p}_i}\\
    &= H(X) + \sum_i p_i \log \frac{p_i}{\tilde{p}_i}.
\end{align}
The sum \(\sum_i p_i \log (p_i/\tilde{p}_i)\) is an example of a Kullback-Leibler
divergence (KL divergence). It can be shown, using Jensen's inequality, that it
is non-negative, and equal to zero only when \(p_i = \tilde{p}_i\) for all
\(i\), i.e.\ only when the model perfectly fits the true generating
distribution (see \citealp{mackay2003} for more detail). For an in-depth
overview of explicit probabilistic model-based approaches to lossless
compression, see \citet{steinruecken2014b}.

\section{Basic coding techniques}\label{sec:old-coding-techniques}
We now give an overview of algorithms for lossless compression and the
compression rates which they achieve. For Huffman coding, arithmetic coding
(AC) and dictionary coding we give only brief overviews, since all of these
algorithms are already described in many textbooks including
\citet{mackay2003,cover1991}. The more recently invented asymmetric numeral
systems (ANS) is introduced briefly here and described in detail in
\Cref{chap:ans}.

\subsection{Huffman coding}
Huffman coding \citep{huffman1952} is a `symbol code', which means it directly
maps individual symbols, from an alphabet \(\mathcal{A}_X\), to binary words.
Huffman coding assumes access to a model ensemble over symbols, i.e.\ a set of
probabilities \(\tilde{p}_1, \ldots, \tilde{p}_I\). A string containing
concatenated Huffman code words can be unambiguously decoded because the words
which the Huffman coder outputs have the `prefix property', which means that no
binary code word is equal to the start of any other code word. For example, it
is possible for the codewords \texttt{0} and \texttt{10} to be output by a
Huffman coder, but then the codeword \texttt{010} would not be permitted,
because it would be impossible to disambiguate this word from the concatenation
of the two words \texttt{0} and \texttt{10}.

Given an ensemble \(X\), Huffman coding uses an efficient iterative algorithm to
generate codewords with length \(l_i\) satisfying
\begin{equation}\label{eq:huff-bound}
  l_i = \ceil{h(x=a_i;\theta)} < h(x=a_i;\theta) + 1,
\end{equation}
where \(\ceil{x}\) denotes the nearest integer greater than or equal to \(x\).
The average length is therefore equal to the entropy \(H(X)\) when: (a) the
model is equal to the true generating distribution and (b) all of the
probabilities are exact integer powers of \(2\), i.e.\ when each \(p_i =
2^{-n_i}\) for \(n_i \in \{0, 1, \ldots\}\). A Huffman code is optimal within
the class of symbol codes (codes that directly map symbols to code words)
assuming only condition (a); i.e.\ even when the average codeword length is
not equal to the entropy, it is not possible to do any better than
Huffman, this is a consequence of the fact that codewords lengths must be whole
numbers, and the Huffman lengths are the smallest integers greater than or
equal to the information content \citep{mackay2003}.

Despite achieving optimal compression rates for individual symbols, Huffman
coding can be highly inefficient for compressing \(sequences\) of symbols,
particularly when the information content of each individual symbol in a
sequence is close to zero, when the contribution from the `\(+\:1\)' in the
upper bound in \cref{eq:huff-bound} can dominate, leading to a total average
message length that is much larger than the sequence's entropy. This
inefficiency is addressed by stream codes, which achieve near-optimal
per-symbol compression rates over sequences of symbols, rather than for
single symbols.

\subsection{Stream codes}
There are two broad classes of codes which work on sequential or streaming
data. The first, which we will refer to as `model-based stream codes', are, in
a sense, a direct generalization of Huffman codes to sequential data. Like
Huffman codes, they require access to a model over data, but unlike Huffman
codes they cater particularly to auto-regressive (sometimes referred to as
`adaptive') models over sequences of symbols. To be precise, for data drawn
from a sequence \(X_1, \ldots, X_N\), they assume access to the CDFs and
inverse CDFs under  the conditional distributions \(P(x_1), P(x_2\given
x_1),\ldots,P(x_N\given x_1,\ldots,x_{N-1})\).

The second major class of stream codes used in practice are known as
`dictionary codes'; they require no prior knowledge, or model, of the
distribution of the input stream and aim to achieve acceptable (as opposed to
optimal) performance for \emph{any} input.

\subsubsection{Arithmetic coding and asymmetric numeral systems}
`Arithmetic coding' \citep[AC;][]{witten1987} and the much more recent
`asymmetric numeral systems' \citep[ANS;][]{duda2009} are both model-based
stream codes, which work by encoding a sequence of symbols one-at-a-time, and
differ from each other in the order in which data may be decoded.  AC is
first-in-first-out (FIFO) or `queue-like', so data are decoded in the
\emph{same} order in which they are encoded, whilst ANS is last-in-first-out
(LIFO) or `stack-like', which means that in ANS, data are decoded in the
\emph{opposite} order to that in which they were encoded. Both schemes have
very similar compression performance, achieving per-symbol compression rates
which are close to the information content under the model.  \citet{mackay2003}
shows that AC can achieve a message length with the worst-case bound
\begin{equation}\label{eq:ac-bound}
  l(x)\stackrel{\mathrm{approx}}{\leq} h(x;\theta) + 2,
\end{equation}
where \(h(x;\theta)\) denotes the information content of an entire sequence,
which can be decomposed as
\begin{equation}
  h(x;\theta) = \sum_{n=1}^N h(x_n\given x_1,\ldots,x_{n-1};\theta).
\end{equation}
The conditional information contents \(h(x_n\given x_1,\ldots,x_{n-1};\theta)\)
are defined in the obvious way as the information content of \(x_n\) under
the conditional distribution \(P(x_n\given x_1,\ldots,x_{n-1})\). For long
sequences, the effect of the `\(+\:2\)' on the per-symbol compression rate
becomes negligible, hence we can say that the per-symbol compression rate of AC
is close to optimal.

The `\(\mathrm{approx}\)' above the `\(\leq\)' symbol is there because Mackay,
and other works which mention this bound, such as \citet{witten1987} and
\citet{moffat1998}, assume the implementation can use exact (i.e.\ unbounded
precision) rational numbers. With exact arithmetic, AC encode and decode
compute time scales poorly with the sequence length \(N\). In practical
settings it is important to achieve fast runtimes, and thus implementations of
AC which are used in production invariably use faster, fixed precision
arithmetic to approximate the exact algorithm. Another difference between
practical implementations and the ideal implementation to which
\cref{eq:ac-bound} applies is that files are usually comprised of a
whole-number of bytes. We can easily account for this and make the bound more
realistic by rounding the quantity on the right-hand-side of \cref{eq:ac-bound}
up to the nearest multiple of 8:
\begin{equation}\label{eq:ac-bound2}
l(x)\stackrel{\mathrm{approx}}{\leq} \ceil{h(x;\theta) + 2}_8
  \leq h(x;\theta) + 10,
\end{equation}
where we use \(\ceil{x}_n\) to denote the smallest multiple of \(n\) which is
greater than or equal to \(x\).  I have been unable to find analysis of the
worse-case behavior of AC when using more realistic bounded-precision
arithmetic, but it is likely that \cref{eq:ac-bound2} is a reasonably good
approximation in most practical settings.

On the other hand, establishing exact worst-case bounds for practical
implementations of ANS, the algorithm that we use throughout this work, is
fairly straightforward, and in \Cref{chap:ans} we show that under our ANS
implementation we have
\begin{equation}
  l(x)\leq h(x) + N\epsilon + C,
\end{equation}
where \(\epsilon\) and \(C\) are both implementation dependent constants. A
typical setup in our experiments gives \(\epsilon\approx 2.2\times10^{-5}\)
and \(C=64\). Although this bound appears worse than the AC bound in
\cref{eq:ac-bound2}, for long enough sequences the difference has little
practical implication---the `one-off' constant \(C\) has a negligible effect
on the per-symbol compression rate, and \(\epsilon\) amounts to a worst-case
overhead of one bit every \(2.2\times10^5\) symbols. The expected (as opposed
to worst-case) behaviour of ANS has not been exactly characterized but
empirically it is usually significantly better than the above bound
\citep{duda2009}.

In practice, ANS is usually slightly faster than AC, is more straightforward to
implement and is also easier to generalize to a vectorized implementation
\citep{duda2009,giesen2014}. Moreover, the LIFO property of ANS is critical for
the new methods introduced in this thesis. In \Cref{chap:ans} we describe ANS
in detail, and we give a full working Python implementation, which is
only 50 lines in length, in \Cref{app:ans-example}. We discuss vectorization,
which we have also implemented, in \Cref{chap:vectorizing}. A high level
description of AC can be found in \citet{mackay2003}, Chapter 6, and a detailed
description with a full, clear C implementation in \citet{witten1987}.

\subsubsection{Dictionary coding}
Dictionary coding, also referred to as substitution coding, works by going
through a stream of symbols and replacing (`substituting') any sub-sequence
which has already occurred in the stream with a pointer to the sub-sequence's
previous occurrence. Compression is achieved when sequences are repeated whose
length exceeds that of a pointer. The family of Lempel-Ziv coders use this
technique, and a variant called DEFLATE is used in \texttt{gzip}, which is
perhaps the most widely used compression software in existence. Although they
do not make use of a model over data, Lempel-Ziv codes can be shown to be
asymptotically optimal \citep{cover1991}. However, for practical sources and
finite sequences, the lengths of encoded messages are often significantly
greater than the entropy. See \citet{mackay2003}, Section 6.4 for more detail,
including examples of sources on which Lempel-Ziv coding performs poorly.

\section{Image compression in practice}\label{sec:practical-image-compression}
In \Cref{chap:plc,chap:hilloc} we present the BB-ANS algorithm, which extends
ANS to latent variable models. The algorithm is generic, in the sense that it
can be applied with a wide range of latent variable models, over any kind of
data. We chose to demonstrate the method on images because deep generative
models of images are well studied and reasonably straightforward to setup and
train, and because image compression is extremely widely used and well studied,
with good baselines to compare to.

For most image compression use cases, such as communication of images across
the internet and local storage of a photo library, some loss of data is
acceptable, particularly if this can be achieved without affecting the
perceptual quality of the image (i.e.\ the appearance of the image `to the
human eye'). JPEG \citep{wallace1991} has been the dominant lossy format for
compressing photographic images since its introduction in 1992, and is
extremely widely used on the web and as a storage format.

\emph{Lossless} image compression is useful when it is not known in advance
that degradation in an image's quality will be acceptable to the intended
recipient. When using lossy compression there is always a somewhat speculative
decision that needs to be made about how much compression is appropriate. As
argued in \citet{sneyers2016}, it is useful to have a codec that users don't
need to think about before using, and lossless codecs fall into this category.
Moreover, lossless image codecs are typically not specialised to photographs
(as JPEG is), and aim to achieve reasonable performance on any image that a
human might wish to communicate or store.

Lossless compression is also particularly useful for saving intermediate
versions during image editing, where repeatedly editing and saving in a lossy
format would lead to successively more severe degradation of image quality. It
is also especially desirable for archival storage, and for images which are
used in scientific or medical applications; indeed in some jurisdictions it is
illegal for medical images to be stored in a lossy format \citep{liu2017}.

\subsection{Comparing PNG, FLIF, and our new codecs}
We now discuss two existing formats, and how the approach they take differs
from ours. The first is `portable network graphics' (PNG), which is the most
widely used lossless image format today, and is particularly popular for
compressing graphics for the web. PNG development began in the mid 1990s, and
the format was intended to be a more flexible replacement for GIF (another
lossless format), and to be completely unencumbered by patents, which GIF, at
the time, was not. The authors of the format claimed at the time that PNG
compression was ``among the best that can be had without losing image data and
without paying patent or other licensing fees'' \citep{roelofs1999}.

The reference PNG implementation, \texttt{libpng}, is free software, and the
PNG standard and \texttt{libpng} were developed by a small community which
overlapped with the group who developed \texttt{gzip}. PNG compression has two
stages, which are referred to as `preprocessing' and `compression'. The
compression stage takes the result of preprocessing and compresses it using the
same algorithm (and in \texttt{libpng} the same implementation) as
\texttt{gzip}. Thus the job of preprocessing is to reversibly convert the image
to a more `\texttt{gzip} friendly' form.

The preprocessing in PNG consists of passing one of five convolutional filters
over each line of the image. The filters all subtract the values of previous
pixels from the current pixel's value, exploiting the approximate smoothness of
images to try to create a result in which all pixel values are close to zero.
The result of preprocessing, along with a list specifying which of the five
filters was used for each row, is passed forwards to the compression stage,
which uses the same DEFLATE algorithm as \texttt{gzip}, a form of dictionary
coding \citep{roelofs1999}. The method for selecting which filter should be
used for each row is not specified by the PNG standard, but the implementation
in \texttt{libpng} uses a simple heuristic, simply selecting the filter whose
output is closest to zero (in the sense that the sum of the absolute values of
each element in the row is smallest).

\begin{figure}[ht]
\centering
\includegraphics[width=0.7\textwidth]{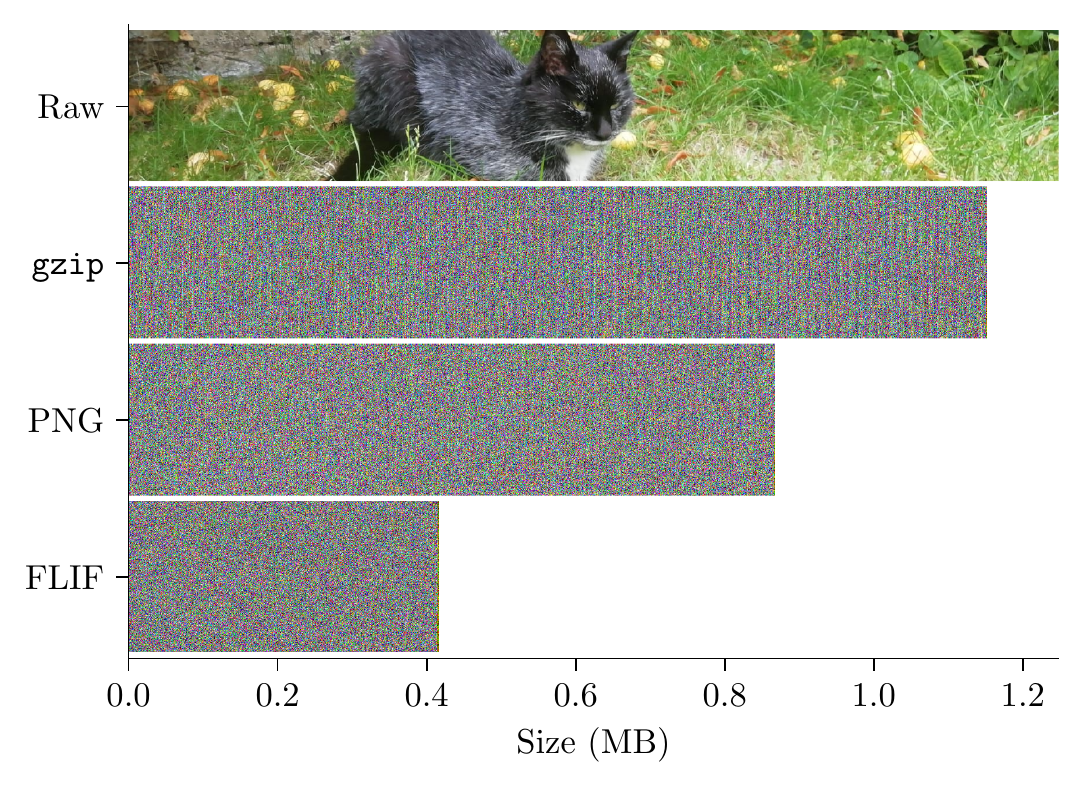}
\caption{
  Comparing a photograph of the cat `Stasha' with the compressed bit streams
  resulting from applying the codecs \texttt{gzip}, PNG and FLIF to the
  image. Vertical bands are clearly visible in the \texttt{gzip} stream, and
  upon close inspection there are some visible vertical bands in the PNG
  stream, implying some correlation between pixels and hence redundancy. No
  structure is visible in the FLIF stream, suggesting that FLIF may be close to
  optimal.
}\label{fig:stasha}
\end{figure}

\Cref{fig:stasha} shows the importance of the preprocessing used by PNG, since
PNG is 23\% smaller than \texttt{gzip} in this case. Displaying a bitstream in
this way is a crude but useful way to observe when a stream contains obvious
redundancy, implying that there is room for improvement in the codec. A perfect
codec should output bits which are indistinguishable from samples from an
i.i.d.\ uniform distribution \citep{mackay2003}. It's impossible to confirm that
a stream is i.i.d.\ uniform using this method, but sometimes non-uniformities
and correlation are obvious enough to be confident that a stream is \emph{not}
i.i.d.\ uniform.

The second format which we discuss is the `free lossless image format' (FLIF),
which was the last to set new state of the art performance on lossless
compression benchmarks, improving on PNG by 43\% on average
\citep{sneyers2016}. FLIF is also free software, though it has yet to gain
widespread adoption\footnote{In fact FLIF has recently been subsumed by the
JPEG XL standard, still under development at time of writing, see
\cite{alakuijala2019}.}.  It is significantly more sophisticated than PNG,
although it also uses a two stage preprocessing and compression pipeline. In
the FLIF documentation, preprocessing is referred to as `prediction', but for
self-consistency we will continue to use `preprocessing'.

Preprocessing in FLIF is actually slightly simpler than in PNG, in that the
same filter is used on every line of the image, instead of switching between
five filters. The filter works by subtracting the median of \(T\), \(L\) and
\(T+L-TL\) from each pixel, where \(T\) is the value of the pixel above, and
\(L\) is the value of the pixel to the left of the current pixel. The
`compression' stage uses an algorithm called `meta-adaptive near-zero integer
arithmetic coding' (MANIAC). MANIAC uses arithmetic coding with an adaptive
model, which learns as it processes an image. Which mixture component is
selected for each pixel is determined by the values of the pixels \(T\) and
\(L\), see \citet{sneyers2016} for more detail.\looseness=-1

\subsubsection{A paradigm shift}
The approach we take to lossless compression is fundamentally different to PNG,
FLIF, and other existing codecs. We explicitly de-couple compression and
modelling, spending a lot of engineering effort and compute time to develop an
accurate image model, \emph{before} doing compression with the model. We can
train our model on a dataset containing a representative sample of the images
that the compression algorithm will be used for. For the full-size colour image
compression experiments in \Cref{chap:hilloc}, we use the ImageNet dataset,
which is reasonably representative of the images that are communicated over the
internet.  Having `seen' a huge number of images before attempting any
compression, our method could be said, in the language of Bayesian statistics,
to have a `strong prior'. Before compressing a test image, it may have very
high level structural knowledge about the kinds of objects that tend to appear
in images. Given an image whose top few rows look like the top of a cat's head,
the model may be able to infer that the rest of the image is likely to contain
the rest of the cat. PNG, FLIF and other existing codecs could only be said to
have weak knowledge about images---they know that pixels are locally correlated
and that patterns tend to be repeated within individual images. They certainly
would not be able to predict the existence of pixels representing a cat's body
given only the top of the cat's head.

This approach represents a paradigm shift, and in \Cref{chap:plc,chap:hilloc}
we demonstrate that this may lead to improved compression rates. Currently, our
implementations are significantly slower than existing codecs, and the BB-ANS
method only performs optimally when processing \emph{batches} of images, but we
are confident that these issues can be overcome, in future work which builds on
the foundation we have helped to lay.

%% file: ANS.tex
\chapter{Introduction to asymmetric numeral systems}\label{chap:ans}

  We are interested in algorithms for lossless compression of sequential data.
  Arithmetic coding (AC) and the range variant of asymmetric numeral systems
  (sometimes abbreviated to rANS, we simply use ANS) are examples of such
  algorithms. Just like arithmetic coding, ANS is close to optimal in terms of
  compression rate \citep{witten1987, duda2009}. The key difference between ANS
  and AC is in the order in which data are \emph{decoded}: in ANS, compression
  is last-in-first-out (LIFO), or `stack-like', while in AC it is
  first-in-first-out (FIFO), or `queue-like'. The stack-like nature of ANS is
  critical for the algorithm we present in \Cref{chap:plc}.  We recommend
  \citet{mackay2003} Chapter 4-6 for background on source coding and arithmetic
  coding in particular. This chapter contains pseudocode which could be
  converted to a working ANS implementation without too much difficulty. We
  also provide a 50 line Python implementation, with example usage, in
  \Cref{app:ans-example}.

  ANS comprises two basic functions, which we denote \push\ and \pop, for
  encoding and decoding, respectively (the names refer to the analogous stack
  operations). The \push\ function accepts some pre-compressed information
  \(m\) (short for `message'), and a symbol \(x\) to be compressed, and returns
  a new compressed message, \(m'\). Thus it has the signature
  \begin{equation}
    \push\colon(m, x) \mapsto m'.
  \end{equation}
  The new compressed message, \(m'\), contains precisely the same information
  as the pair \((m, x)\), and therefore \push\ can be inverted to form a
  decoder mapping.  The decoder, \pop, maps from \(m'\) back to \(m, x\):
  \begin{equation}
    \pop\colon m' \mapsto (m, x).
  \end{equation}
  Because the functions \push\ and \pop\ are inverse to one another, we have
  \(\push(\pop(m))=m\) and \(\pop(\push(m, x)) = (m, x)\).

\section{Specifying the problem which ANS solves}\label{sec:prob-spec}
  In this section we first define some notation, then describe the problem
  which ANS solves in more detail and sketch the high level approach to solving
  it. In the following we use `\(\log\)' as shorthand for the base 2 logarithm,
  usually denoted `\(\log_2\)'.

  The functions \push\ and \pop\ will both require access to the probability
  distribution from which symbols are drawn (or an approximation thereof). To
  describe distributions we use notation similar to \citet{mackay2003}:
  \begin{definition}\label{def:ensemble2}
    A \emph{quantized ensemble} \(X\) with precision \(r\) is a triple \((x,
    \mathcal{A}_X, \mathcal{P}_X)\) where the \emph{outcome} \(x\) is the value
    of a random variable, taking on one of a set of possible values
    \(\mathcal{A}_X = \{a_1, \ldots, a_I\}\), and \(\mathcal{P}_X = \{p_1,
    \ldots, p_I\}\) are the \emph{integer-valued} probability weights with each
    \(p_i\in\{1, \ldots, 2^r\}\), each \(P(x=a_i) = p_i / 2^r\) and therefore
    \(\sum_{i=1}^Ip_i = 2^r\).
  \end{definition}
  Note that this definition differs from \Cref{def:ensemble1} from the last
  chapter, in that the probabilities are assumed to be \emph{quantized}  to
  some precision \(r\) (i.e.\ representable by fractions \(p_i/2^r\)), and we
  assume that none of the \(a_i\) have zero probability. Having probabilities
  in this form is necessary for the arithmetic operations involved in ANS (as
  well as AC). Note that if we use a high enough \(r\) then we can specify
  probabilities that are not close to zero with a precision similar to that of
  typical floating point---32-bit floating point numbers for example contain 23
  `fraction' bits, and thus would have roughly the same precision as our
  representation with \(r=23\). Symbols with very small probabilities are a
  potential failure mode of both ANS and AC. These must either be rounded up to
  the smallest nonzero quantized value, or all symbols with small probabilities
  can effectively be split into two symbols: the first, a global `escape'
  symbol, which has mass equal to the sum of the small probabilities, and the
  second, the symbol itself, with mass equal to the conditional given that the
  symbol is in the escape group (this conditional will be larger than the
  symbol's original mass). An escape mechanism is described in more detail in
  \citet{moffat1998}.

  Given a sequence of quantized ensembles \(X_1, \ldots, X_N\), we seek an
  algorithm which can encode any outcome \(x_1, \ldots, x_N\) in a binary
  message whose length is close to \(h(x_1, \ldots, x_N) = \log 1/P(x_1,
  \ldots, x_N)\).  According to Shannon's source coding theorem it is not
  possible to losslessly encode data in a message with expected length less
  than \(\mathbb{E}[h(x)]\), thus we are looking for an encoding which is close
  to optimal in expectation \citep{shannon1948}. Note that the joint
  information content of the sequence can be decomposed:
  \begin{align}
    \label{eq:info-decomp1}
    h(x_1, \ldots, x_N)
      &= \log\frac{1}{P(x_1, \ldots, x_N)}\\
      \label{eq:info-decomp2}
      &= \sum_n \log\frac{1}{P(x_n \given x_1, \ldots, x_{n-1})}\\
      \label{eq:info-decomp3}
      &= \sum_n h(x_n \given x_1, \ldots, x_{n-1}).
  \end{align}

  Because it simplifies the presentation significantly, we focus first on the
  ANS \emph{decoder}, the reverse mapping which maps from a compressed binary
  message to the sequence \(x_1, \ldots, x_N\). This will be formed of a
  sequence of \(N\) \pop\ operations; starting with a message \(m_0\) we define
  \begin{align}
    m_n, x_n = \pop(m_{n-1})\qquad\text{for }n=1, \ldots, N
  \end{align}
  where each \pop\ uses the conditional distribution \(X_n\given X_1, \ldots,
  X_{n-1}\). We will show that the message resulting from each \pop, \(m_n\),
  is effectively shorter than \(m_{n-1}\) by no more than \(h(x_n \given x_1,
  \ldots, x_{n-1}) + \epsilon\) bits, where \(\epsilon\) is a small constant
  which we specify below, and therefore the difference in length between
  \(m_0\) and \(m_N\) is no more than \(h(x_1, \ldots, x_N) + N\epsilon\), by
  \cref{eq:info-decomp1,eq:info-decomp2,eq:info-decomp3}.

  We will also show that \pop\ is a bijection whose inverse, \push, is
  straightforward to compute, and therefore an encoding procedure can easily be
  defined by starting with a very short base message and adding data
  sequentially using \push. Our guarantee about the effect of \pop\ on message
  length translates directly to a guarantee about the effect of \push, in that
  the increase in message length due to the sequence of \push\ operations is
  less than \(h(x_1, \ldots, x_N) + N\epsilon\).

\section{Asymmetric numeral systems}\label{sec:ans-ans}
  Having set out the problem which ANS solves and given a high level overview
  of the solution in \Cref{sec:prob-spec}, we now go into more detail, firstly
  discussing the data structure we use for \(m\), then the \pop\ function and
  finally the computation of its inverse, \push.

\subsection{The structure of the message}\label{sec:message}
\begin{figure}[ht]
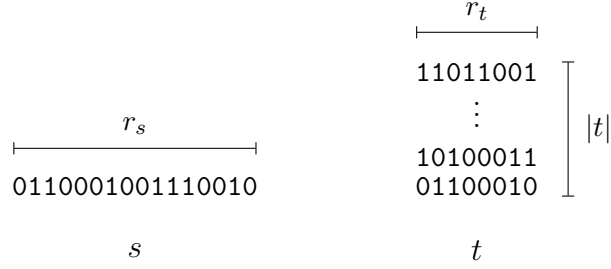

    \centering
    \drawmessage
    \caption{
      The two components of a message: the unsigned integer \(s\) (with \(r_s =
      16\)) and the stack of unsigned integers \(t\) (with \(r_t = 8\)). The
      integers are represented here in base 2 (binary).
  }
    \label{fig:message}
  \end{figure}

  We use a pair \(m = (s, t)\) as the data structure for the message \(m\). The
  element \(s\) is an unsigned integer with precision \(r_s\) (i.e.\ \(s \in
  \{0, 1, \ldots, 2^{r_s} - 1\}\), so that \(s\) can be expressed as a binary
  number with \(r_s\) bits). The element \(t\) is a stack of unsigned
  integers of some fixed precision \(r_t\) where \(r_t < r_s\).  This stack has
  its own push and pop operations, which we denote \(\texttt{stack\_push}\) and
  \(\texttt{stack\_pop}\) respectively. See \cref{fig:message} for a diagram of
  \(s\) and \(t\).  We need \(s\) to be large enough to ensure that our
  decoding is accurate, and so we also impose the constraint
  \begin{equation}\label{eq:s-constraint}
    s\geq2^{r_s - r_t},
  \end{equation}
  more detail on how and why we do this is given below. In the demo
  implementation we use \(r_s = 64\) and \(r_t = 32\).

  Note that a message can be flattened into a string of bits by concatenating
  \(s\) and the elements of \(t\). The length of this string is
  \begin{equation}
    l(m) := r_s + r_t\abs{t},
  \end{equation}
  where \(\abs{t}\) is the number of elements in the stack \(t\). We refer to
  this quantity as the `length' of \(m\). We also define the useful quantity
  \begin{equation}
    l^*(m) := \log s + r_t\abs{t},
  \end{equation}
  which we refer to as the `effective length' of \(m\). Note that the
  constraint in \cref{eq:s-constraint} and the fact that \(s < 2^{r_s}\) imply
  that
  \begin{equation}\label{eq:effective-length}
    l(m) - r_t \leq l^*(m) < l(m).
  \end{equation}
  Intuitively \(l^*\) can be thought of as a precise measure of the size of
  \(m\), whereas \(l\), which is integer valued, is a more crude measure.
  Clearly \(l\) is ultimately the measure that we care most about, since it
  tells us the size of a binary encoding of \(m\), and we use \(l^*\) to prove
  bounds on \(l\).

\subsection{Constructing the pop operation}\label{sec:scalar-pop}
  To avoid notational clutter, we begin by describing the \pop\ operation for a
  single quantized ensemble \(X = (x, \mathcal{A}_X, \mathcal{P}_X)\) with
  precision \(r\), before applying \pop\ to a sequence in \Cref{seq:pop-seq}.
  Our strategy for performing a decode with \pop\ will be firstly to extract a
  symbol from \(s\). We do this using a bijective function \(d\colon\mathbb
  N\rightarrow\mathbb N\times\mathcal{A}\), which takes an integer \(s\) as
  input and returns a pair \((s', x)\), where \(s'\) is an integer and \(x\) is
  a symbol. Thus \pop\ begins
  \begin{singlespace}
  \begin{lstlisting}
def pop($m$):
    $s$, $t$ := $m$
    $s'$, $x$ := $d(s)$
  \end{lstlisting}
  \end{singlespace}
We design the function \(d\) so that if \(s\geq 2^{r_s - r_t}\), then
\begin{equation}\label{eq:inner-decoder}
  \log s - \log s' \leq h(x) + \epsilon
\end{equation}
where
\begin{equation}
  \epsilon := \log \frac{1}{1 - 2^{-(r_s - r_t - r)}}.
\end{equation}
We give details of \(d\) and prove \cref{eq:inner-decoder} below. Note that
when the term \(2^{-(r_s - r_t - r)}\) is small, the following approximation
is accurate:
  \begin{equation}
    \epsilon \approx \frac{2^{-(r_s - r_t - r)}}{\ln 2},
  \end{equation}
  and thus $\epsilon$ itself is small. We typically use \(r_s = 64\), \(r_t =
  32\), and \(r = 16\), which gives \(\epsilon = \log 1/(1 - 2^{-16}) \approx
  2.2 \times 10^{-5}\).

  After extracting a symbol using \(d\), we check whether \(s'\) is below
  \(2^{r_s - r_t}\), and if it is we \texttt{stack\_pop} integers from \(t\)
  and move their contents into the lower order bits of \(s'\). We refer to this
  as `renormalization'.  Having done this, we return the new message and the
  symbol \(x\). The full definition of \pop\ is thus
  \begin{singlespace}
  \begin{lstlisting}[frame=single]
def pop($m$):
    $s$, $t$ := $m$
    $s'$, $x$ := $d$($s$)
    $s$, $t$ := renorm($s'$, $t$)
    return ($s$, $t$), $x$
  \end{lstlisting}
  \end{singlespace}

  Renormalization is necessary to ensure that the value of \(s\) returned by
  \pop\ satisfies \(s\geq2^{r_s - r_t}\) and is therefore large enough that
  \cref{eq:inner-decoder} holds at the start of any future \pop\ operation. The
  \texttt{renorm} function has a while loop, which pushes elements from \(t\)
  into the lower order bits of \(s\) until \(s\) is full to capacity. To be
  precise:
  \begin{singlespace}
  \begin{lstlisting}[frame=single]
def renorm($s$, $t$):
    # while $s$ has space for another element from $t$
    while $s < 2^{r_s - r_t}$:
        # pop an element $t_\mathrm{top}$ from $t$
        $t$, $t_{\mathrm{top}}$ := stack_pop($t$)
        # and push $t_\mathrm{top}$ into the lower bits of $s$
        $s$ := $2^{r_t} \cdot s$ + $t_{\mathrm{top}}$
    return $s$, $t$
  \end{lstlisting}
  \end{singlespace}

  The condition \(s < 2^{r_s - r_t}\) guarantees that \(2^{r_t} \cdot s +
  t_{\text{top}} < 2^{r_s}\), and thus there can be no loss of information
  resulting from overflow. We also have
  \begin{equation}
    \log (2^{r_t} \cdot s + t_\text{top}) \geq r_t + \log s
  \end{equation}
  since \(t_{\text{top}} \geq 0\).  Applying this inequality repeatedly, once
  for each iteration of the while loop in \texttt{renorm}, we have
  \begin{equation}\label{eq:renorm}
    \log s \geq \log s' + r_t\cdot\left[\text{\# elements popped from
    \(t\)}\right],
  \end{equation}
  where \(s, t = \texttt{renorm}(s', t)\) as in the definition of \pop.

  Combining \cref{eq:inner-decoder} and \cref{eq:renorm} gives us
  \begin{equation}\label{eq:pop-inequality}
    l^*(m) - l^*(m') \leq h(x) + \epsilon,
  \end{equation}
  where \((m', x) = \pop(m)\), using the definition of \(l^*\). That is, the
  reduction in the effective message length resulting from \pop\ is close to
  \(h(x)\).

\subsection{Popping in sequence}\label{seq:pop-seq}
  We now apply \pop\ to the setup described in \Cref{sec:prob-spec}, performing
  a sequence of \pop\ operations to decode a sequence of data. We suppose that
  we are given some initial message \(m_0\).

  For \(n=1\ldots N\), we let \(m_n, x_n = \pop(m_{n-1})\) as in
  \Cref{sec:prob-spec}, where each \pop\ uses the corresponding distribution
  \(X_n \given X_1, \ldots, X_{n-1}\). Applying \cref{eq:pop-inequality} to
  each of the \(N\) \pop\ operations, we have:
  \begin{align}
    l^*(m_0) - l^*(m_N)
      &= \sum_{n=1}^N [l^*(m_{n-1}) - l^*(m_n)]\\
      &\leq \sum_{n=1}^N [h(x_n \given x_1, \ldots, x_{n-1}) + \epsilon]\\
      &\leq h(x_1, \ldots, x_N) + N\epsilon.\label{eq:effective-lengths}
  \end{align}

  This result tells us about the reduction in message length from \pop\, but
  also, conversely, about the length of a message \emph{constructed} using
  \push. We can actually initialize an encoding procedure by \emph{choosing}
  \(m_N\), and then performing a sequence of \push\ operations. Since our
  ultimate goal when encoding is to minimize the encoded message length \(m_0\)
  we choose the setting of \(m_N\) which minimizes \(l^*(m_N)\), which is \(m_N
  = (s_N, t_N)\) where \(s_N = 2^{r_s - r_t}\) and \(t_N\) is an empty stack.
  That gives \(l^*(m_N) = r_s - r_t\) and therefore, by
  \cref{eq:effective-lengths},
  \begin{equation}
    l^*(m_0) \leq h(x_1, \ldots, x_N) + N\epsilon + r_s - r_t.
  \end{equation}
  Combining that with \cref{eq:effective-length} gives an expression for the
  actual length of the flattened binary message resulting from \(m_0\):
  \begin{equation}\label{eq:length-bound}
    l(m_0) \leq h(x_1, \ldots, x_N) + N\epsilon + r_t.
  \end{equation}

  It now remains for us to describe the function \(d\) and show that it
  satisfies \cref{eq:inner-decoder}, as well as showing how to invert \pop\ to
  form the encoding function \push.

\subsection{The function \(d\)}
The function \(d\colon\mathbb{N}\rightarrow\mathbb{N}\times\mathcal{A}\) must be a
bijection, and we aim for \(d\) to satisfy \cref{eq:inner-decoder}, and thus
\(P(x)\approx \frac{s'}{s}\).  Achieving this is actually fairly
straightforward.  One way to define a bijection \(d\colon s\mapsto(s', x)\) is
to start with a mapping \(\tilde d\colon s\mapsto x\), with the property that
none of the preimages \(\tilde d^{-1}(x):=\{n\in\mathbb{N}:\tilde d(n) = x\}\)
are finite for \(x\in\mathcal{A}\). Then let \(s'\) be the index of \(s\)
within the (ordered) set \(\tilde d^{-1}(x)\), with indices starting at \(0\).
Equivalently, \(s'\) is the number of integers \(n\) with \(0\leq n<s\) and
\(d(n) = x\).

With this setup, the ratio
\begin{equation}\label{eq:ratio}
  \frac{s'}{s} = \frac{\abs{\{n\in\mathbb{N}: n < s, d(n) = x\}}}{s}
\end{equation}
is the density of numbers which decode to \(x\), within all the natural numbers
less than \(s\). For large \(s\) we can ensure that this ratio is close to
\(P(x)\) by setting \(\tilde d\) such that numbers which decode to a symbol
\(x\) are distributed \emph{within the natural numbers} with density close to
\(P(x)\).

To do this, we partition \(\mathbb{N}\) into finite ranges of equal length, and
treat each range as a model for the interval \([0, 1]\), with sub-intervals
within \([0, 1]\) corresponding to each symbol, and the width of each
sub-interval being equal to the corresponding symbol's probability (see
\cref{fig:interval}). To be precise, the mapping \(\tilde d\) can then be
expressed as a composition \(\tilde d = \tilde d_2 \circ \tilde d_1\), where
\(\tilde d_1\) does the partitioning described above, and \(\tilde d_2\)
assigns numbers within each partition to symbols (sub-intervals). So
\begin{equation}
  \tilde d_1(s) := s \bmod 2^{r}.
\end{equation}
Using the shorthand \(\bar{s} := \tilde d_1 (s)\), and defining
\begin{equation}
  c_j := \begin{cases}
    0                    &\quad\text{if }j=1\\
    \sum_{k=1}^{j-1} p_k &\quad\text{if }j=2,\ldots,I
  \end{cases}
\end{equation}
as the (quantized) cumulative probability of symbol \(a_{j-1}\),
\begin{equation}
  \tilde d_2(\bar s) := a_i\text{ where }i := \max \{j : c_j \leq \bar s\}.
\end{equation}
That is, \(\tilde d_2(\bar s)\) selects the symbol whose sub-interval contains
\(\bar s\).  \Cref{fig:interval} illustrates this mapping, with a particular
probability distribution, for the range \(s = 64,\ldots, 71\).

\begin{figure}[ht]
  \centering
  \drawinterval
  \caption{
    Showing the correspondence between \(s\), \(s \bmod 2^{r}\) and the
    symbol \(x\). The interval \([0, 1]\subset\mathbb{R}\) is modelled by the
    set of integers \(\{0, 1, \ldots, 2^{r} - 1\}\). In this case \(r = 3\)
    and the probabilities of each symbol are \(P(\mathtt a) =
    \nicefrac{1}{8}\), \(P(\mathtt b) = \nicefrac{2}{8}\), \(P(\mathtt c) =
    \nicefrac{3}{8}\) and \(P(\mathtt d) = \nicefrac{2}{8}\).}
  \label{fig:interval}
\end{figure}

\begin{figure}[ht]
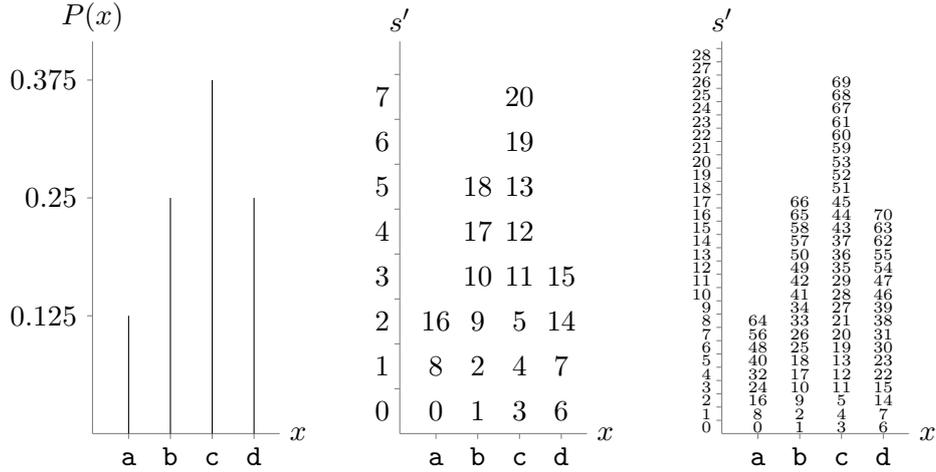

  \centering
  \drawpmf \quad \drawapprox{20}{} \quad \drawapprox{70}{\tiny}
  \caption{
    Showing the pmf of a distribution over symbols (left) and a visualization
    of the mapping \(d\) (middle and right). In the middle and right figures,
    numbers less than or equal to \(s_\mathrm{max}\) are plotted, for
    \(s_\mathrm{max}=20\) and \(s_\mathrm{max}=70\).  The position of each
    number \(s\) plotted is set to the coordinates \((x, s')\), where \(s', x =
    d(s)\). The heights of the bars are thus determined by the ratio \(s'/s\)
    from \cref{eq:ratio}, and can be seen to approach the heights of the lines
    in the histogram on the left (that is, to approach \(P(x)\)) as the density
    of numbers increases.
  }\label{fig:visual-ans}
\end{figure}

\subsection{Computing \(s'\)}
The number \(s'\) was defined above as ``the index of \(s\) within the
(ordered) set \(\tilde d^{-1}(x)\), with indices starting at \(0\)''. We now
derive an expression for \(s'\) in terms of \(s\), \(p_i\) and \(c_i\), where
\(i = \max\{j: c_j \leq \bar s\}\) (as above), and we prove
\cref{eq:inner-decoder}.

Our expression for \(s'\) is a sum of two terms. The first term counts the
entire intervals, corresponding to the selected symbol \(a_i\), which are below
\(s\). The size of each interval is \(p_i\) and the number of intervals is
\(s\div 2^{r}\), thus the first term is \(p_i \cdot (s \div 2^{r})\), where
\(\div\) denotes \emph{integer} division, discarding any remainder.  The second
term counts our position within the current interval, which is \(\bar s - c_i
\equiv s\bmod 2^{r} - c_i\). Thus
\begin{equation}\label{eq:s' def}
  s' = p_i \cdot (s \div 2^{r}) + s\bmod 2^{r} - c_i.
\end{equation}
This expression is straightforward to compute. Moreover from this expression it
is straightforward to prove \cref{eq:inner-decoder}. Firstly, taking the
\(\log\) of both sides of \cref{eq:s' def} and using the fact that \(s\bmod
2^{r} - c_i \geq 0\) gives
\begin{align}
  \log s' \geq \log (p_i\cdot (s\div 2^{r})).
\end{align}
then by the definition of \(\div\), we have \(s\div 2^{r} > \frac{s}{2^{r}}
- 1\), and thus
\begin{align}
  \log s'
    &\geq \log\left(p_i\left(\frac{s}{2^{r}} -1\right)\right)\\
    &\geq \log s - h(x) + \log\left(1 - \frac{2^{r}}{s}\right)\\
    &\geq \log s - h(x) - \epsilon,
\end{align}
as required, using the fact that \(P(x) = \frac{p_i}{2^{r}}\) and \(s \geq
2^{r_s - r_t}\).

By choosing \(r_s - r_t\) to be reasonably
large (it is equal to 32 in our implementation), we ensure that
\(\frac{s'}{s}\) is very close to \(P(x)\). This behaviour can be seen visually
in \cref{fig:visual-ans}, which shows the improvement in the approximation for
larger \(s\).\looseness=-1

\subsection{Pseudocode for \(d\)}
We now have everything we need to write down a procedure to compute \(d\). We
assume access to a function \(f_X\colon \bar{s}\mapsto (a_i, c_i, p_i)\), where
\(i\) is defined above. This function clearly depends on the distribution of
\(X\), and its computational complexity is equivalent to that of computing the
CDF and inverse CDF for \(X\). For many common distributions, the CDF and
inverse CDF have straightforward closed form expressions, which don't require
an explicit sum over \(i\).

We compute \(d\) as follows:
\begin{singlespace}
\begin{lstlisting}[frame=single]
def $d$($s$):
    $\bar s$ := $s\bmod 2^{r}$
    $x$, $c$, $p$ := $f_X(\bar s)$
    $s'$ := $p \cdot (s \div 2^{r}) + \bar s - c$
    return $s'$, $x$
\end{lstlisting}
\end{singlespace}

\subsection{Inverting the decoder}
Having described a decoding process which appears not to throw away any
information, we now derive the inverse process, \push, and show that it is
computationally straightforward.

The \push\ function has access to the symbol \(x\) as one of its inputs, and
must do two things. Firstly it must \texttt{stack\_push} the correct number of
elements to \(t\) from the lower bits of \(s\). Then it must reverse the effect
of \(d\) on \(s\), returning a value of \(s\) identical to that before \pop\
was applied.

Thus, on a high level, the inverse of the function \pop\ can be expressed as
\begin{singlespace}
\begin{lstlisting}[frame=single]
def push($m$, $x$):
    $s$, $t$ := $m$
    $p$, $c$ := $g_X$($x$)
    $s'$, $t$ := renorm_inverse($s$, $t$; $p$)
    $s$ := $d^{-1}$($s'$; $p$, $c$)
    return $s$, $t$
\end{lstlisting}
\end{singlespace}
where \(g_X:x\mapsto (p_i, c_i)\) with \(i\) as above. The function \(g_X\) is
similar to \(f_X\) in that it is analogous to computing the quantized CDF and
mass function \(x \mapsto p_i\).  The function \(d^{-1}\) is really a
pseudo-inverse of \(d\); it is the inverse of \(s\mapsto d(s, x)\), holding
\(x\) fixed.

As mentioned above, \texttt{renorm\_inverse} must \texttt{stack\_push} the
correct amount of data from the lower order bits of \(s\) into \(t\). A
necessary condition which the output of \texttt{renorm\_inverse} must satisfy
is
\begin{equation}\label{eq:renorm_inverse_ineq}
  2^{r_s - r_t} \leq d^{-1}(s'; p, c) < 2^{r_s}.
\end{equation}
This is because the output of \push\ must be a valid message, as described in
\Cref{sec:message}, just as the output of \pop\ must be.

The expression for \(s'\) in \cref{eq:s' def} is straightforward to invert,
yielding a formula for
\(d^{-1}\):
\begin{equation}
  d^{-1}(s'; p, c) = 2^{r} \cdot (s' \div p) + s' \bmod p + c.
\end{equation}
We can substitute this into \cref{eq:renorm_inverse_ineq} and simplify:
\begin{align}
      &&2^{r_s - r_t}            &\leq 2^{r} \cdot (s' \div p) + s' \bmod p + c < 2^{r_s}\\
  \iff&&2^{r_s - r_t}            &\leq 2^{r} \cdot (s' \div p) < 2^{r_s}\\
  \iff&&p\cdot2^{r_s - r_t - r}&\leq s' < p\cdot 2^{r_s - r}.\label{eq:simple-cond}
\end{align}
So \texttt{renorm\_inverse} should move data from the lower order bits of
\(s'\) into \(t\) (decreasing \(s'\)) until \cref{eq:simple-cond} is satisfied.
To be specific:

\begin{singlespace}
\begin{lstlisting}[frame=single]
def renorm_inverse($s'$, $t$; $p$):
    while $s' \geq p \cdot 2^{r_s - r}$:
        $t$ := stack_push($t$, $s'\bmod 2^{r_t}$)
        $s'$ := $s'\div 2^{r_t}$
    return $s'$, $t$
\end{lstlisting}
\end{singlespace}

Although, as mentioned above, \cref{eq:simple-cond} is a \emph{necessary}
condition which \(s'\) must satisfy, it isn't immediately clear that it's
sufficient. Is it possible that we need to continue the while loop in
\texttt{renorm\_inverse} past the first time that \(s'<p\cdot2^{r_s - r}\)? In
fact this can't be the case, because \(s'\div2^{r_t}\) decreases \(s'\) by a
factor of at least \(2^{r_t}\), and thus as we iterate the loop above we will
land in the interval specified by \cref{eq:simple-cond} at most once. This
guarantees that the \(s\) that we recover from \texttt{renorm\_inverse} is the
correct one.

\section{Further reading}
Since its invention by \citet{duda2009}, ANS appears not to have gained
widespread attention in academic literature, despite being used in various
state of the art compression systems. At the time of writing, a search on
Google Scholar for the string ``asymmetric numeral systems'' yields 148 results.
For comparison, a search for ``arithmetic coding'', yields `about 44,000'
results. As far as I'm aware, ANS has appeared in only one textbook, with a
practical, rather than mathematical, presentation \citep{mcanlis2016}.

However, for those wanting to learn more there is a huge amount of material on
different variants of ANS in \citet{duda2009} and \citet{duda2015}. Extensions
to latent variable models are, of course, described in
\Cref{chap:plc,chap:hilloc}, as well as in
\citet{townsend2019,kingma2019,townsend2020}. A parallelized implementation
based on SIMD instructions was first presented in \citet{giesen2014} and is
described in \Cref{chap:vectorizing}. Finally, a version which performs
simultaneous encryption and compression is described in \citet{duda2016}.

Duda maintains a list of ANS implementations at
\url{
  https://encode.su/
  threads/2078-List-of-Asymmetric-Numeral-Systems-implementations}.

%% file: PLC.tex
\chapter{Bits back coding with asymmetric numeral systems}\label{chap:plc}

This chapter is based on the paper `Practical lossless compression with latent
variables using bits back coding' \citep{townsend2019}, which was presented at
the International Conference of Learning Representations (ICLR)\@. The paper
was co-authored with Tom Bird, who assisted with writing the paper and setting
up the experiments. We propose a new coding method, extending ANS to latent
variable models. Like ANS itself, our new method achieves a near-optimal rate
when a sequence (or `batch') of data are compressed, and may suffer a
significant, but bounded, overhead at the beginning of the encoding process (we
discuss methods for minimizing this overhead in \Cref{sec:starting}). At the
end of this chapter we demonstrate the method by compressing the MNIST test set
using a VAE, outperforming all other codecs benchmarked, in terms of
compression rate.  Code for reproducing the results in this chapter can be
found at \jurl{github.com/bits-back/bits-back}.

The lossless compression algorithms mentioned in \Cref{chap:bg,chap:ans}, namely
Huffman coding, arithmetic coding (AC) and asymmetric numeral systems (ANS), do
not naturally cater for models with latent variables. However, there is a
method, known as `bits back coding' \citep{wallace1990,hinton1993}, first
introduced as a thought experiment, but later implemented in \citet{frey1996}
and \citet{frey1997}, which may be used to extend those algorithms to such
models.

Although bits back coding was implemented in restricted cases by
\citet{frey1997}, prior to our work there was no known implementation for modern
neural net-based models or high dimensional data; Frey's implementation was
demonstrated on \(8\times8\) binary images. There is, in fact, an awkward
incompatibility between bits back and the arithmetic coding scheme with which
it was implemented in \citet{frey1997}. In this chapter we describe this issue
and present a clean solution---a scheme that instead implements bits back
coding using ANS. We term this new coding scheme `Bits Back with ANS' (BB-ANS).

In \Cref{mnist_experiments} we demonstrate the efficacy of BB-ANS by
losslessly compressing the MNIST dataset with a variational auto-encoder
\citep[VAE;][]{kingma2014}, a deep latent variable model with continuous latent
variables which we described in \Cref{sec:lvms}. BB-ANS with a VAE outperforms
generic compression algorithms for both binarized and raw MNIST, even with a
very simple model architecture. In \Cref{chap:hilloc} we demonstrate that the
method scales well to a larger model, using it to compress full-size colour
photographs.

\begin{figure}[t]
\includegraphics[width=\textwidth]{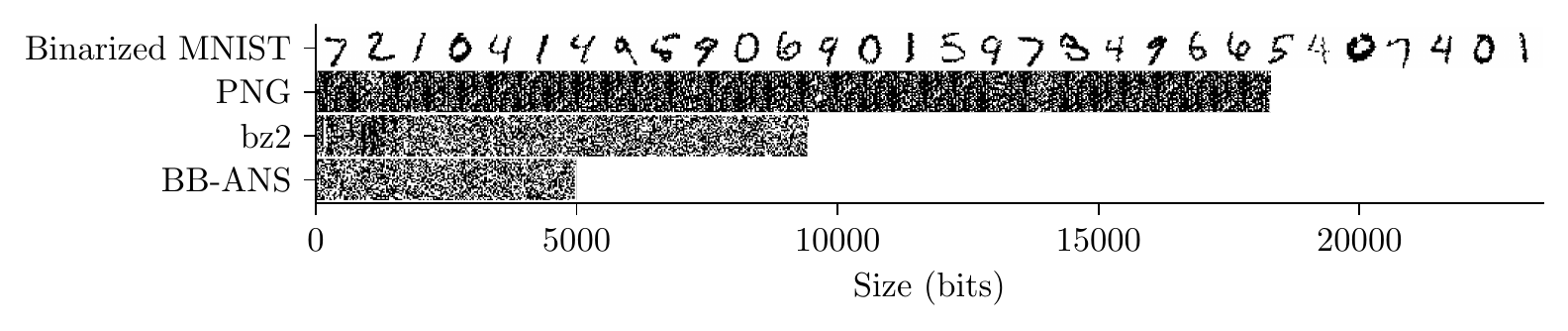}
\caption{Visual comparison of 30 binarized MNIST images with bit stream outputs
from running lossless compression algorithms PNG, bz2 and BB-ANS on the
images.}\label{fig:visual_compression}
\end{figure}
\section{Bits back coding}\label{bbc}
It is well known that an arithmetic \emph{decoder} can be used to map a
randomised binary message to a sample from the distribution used as a model for
the coder.  In the case of ANS, the fact that running the decoder,
parametrized by some mass function \(P\), on a random  message will generate a
sample from \(P\), is a straightforward consequence of the fact, stated and
proved in \Cref{sec:ans-ans}, that numbers which decode to \(x\) appear in the
natural numbers with density close to \(P(x)\). This is a fundamental property
of ANS which is necessary for optimal compression.

In this section we describe bits back coding, a method which uses the
aforementioned sampling capability of a lossless codec for compression of data
using a latent variable model. We first present the form in which bits-back
coding has appeared in previous works, then we present our own novel approach.

\subsection{Bits back without ANS}\label{sec:bbc-sub}
\begin{figure}[h]
\centering
\begin{tikzpicture}
  \tikzstyle{var}=[circle,draw=black,minimum size=7mm]
  \tikzstyle{latent}  =[]
  \tikzstyle{observed}=[fill=gray!50]
  \node at (0,   0) [var,latent]   (z)  {\(z\)};
  \node at (1.5, 0) [var,observed] (x)  {\(x\)};
  \draw[->]  (z) -- (x);
\end{tikzpicture}
\caption{
  Graphical model with latent variable \(z\) and observed variable \(x\).
}
\end{figure}
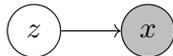
Suppose that a sender wishes to communicate a symbol $x$ to a receiver, and
that both sender and receiver have access to a generative model with a latent
variable, $z$. For now we take $z$ to be discrete; in \Cref{sec:plc-disc} we
show how to apply the method to continuous latents.

Suppose that the CDFs/inverse CDFs required for coding with ANS under the
marginal distribution \(P(x)\) are intractable to compute, as is typically the
case in deep latent variable models. Bits back is an idea that can be used to
encode and decode \(x\) assuming only the ability to encode/decode using the
forward probabilities $P(z)$ and $P(x\given z)$, and a (possibly approximate)
posterior $Q(z\given x)$.

We must assume that, as well as the sample $x$, the sender already has a
separate compressed message to communicate. The sender can run the approximate
posterior \emph{decode} on the extra message to generate a sample from
$Q(z\given x)$. Then they can encode the latent sample according to $P(z)$ and
the symbol $x$ according to $P(x\given z)$. The receiver then does the inverse
to recover the latent sample and the symbol. The extra message can also be
precisely recovered by the receiver by \emph{encoding} the latent sample
according to $Q(z\given x)$.

We can write down the expected increase in message length (over the length of
the extra message at the start):
\begin{align}
  L(Q)
  &= \mathbb{E}_{Q(z\given x)}
    \big[-\log P(z) -\log P(x\given z) + \log Q(z\given x) \big]\\
  &= -\mathbb{E}_{Q(z\given x)}\log\frac{P(x, z)}{Q(z\given x)}.
\end{align}
This is the negative of the evidence lower bound (ELBO), which was defined in
\cref{eq:bg-elbo-def}.

A great deal of recent research has focused on inference and learning with
approximate posteriors, using the ELBO as an objective function. Because of the
above equivalence, methods which maximize the ELBO for a model are implicitly
minimizing the message length achievable by bits back coding with that model.
This suggests that we may be able to draw on this plethora of existing methods
when learning a model for use in compression applications, safe in the
knowledge that the objective function they are maximizing is the negative
expected message length.

\subsection{Chaining bits back coding}\label{sec:plc-chaining}
If we wish to encode a \emph{sequence} of data points, and do not have any extra
information to communicate, we may be able to accept a one-time overhead for
coding the first element at a rate worse than the negative ELBO. Maybe we have
a fallback codec which doesn't require an existing message, which we can use
for the first element; if not, we can generate the latent for the first element
of the sequence in any way we like, including by sampling it from a
pseudo-random number generator, and then encode it using the prior and
likelihood. The resulting compressed message can then be used as the extra
message for bits-back coding of the second data point, the encoded second data
point as the extra message for the third, and so on. This daisy-chain-like
scheme was first described by \citet{frey1997}, and was called `bits-back with
feedback'. We refer to it simply as `chaining'.

As \citet{frey1997} notes, the above method cannot be implemented directly
using AC, because in order for it to work it is necessary to decode data in the
opposite order to that in which they were encoded. Frey gets around this by
implementing what amounts to a stack-like wrapper around AC, for which it is
necessary to terminate AC encoding after each chaining step, effectively using
AC like a symbol code.  This incurs a cost both in code complexity and,
importantly, in compression rate. The cost in compression rate is due to the
fact that terminating AC incurs a cost of up to two bits (see eq.
\ref{eq:ac-bound}). As Frey notes, any symbol code will do for chaining, and in
situations where \(x\) and \(z\) are actually each comprised of individual
symbols (in most situations we study, they are in fact vectors), Huffman coding
would be the optimal choice, but this still incurs a compression rate overhead
for each chaining step.

\subsection{Chaining bits back coding with ANS} \label{sec:bbc_with_ans}
The central insight of this chapter is the observation that the chaining
described in the previous section can be implemented straightforwardly with ANS
with zero compression rate overhead per iteration. This is because of the fact
that ANS is stack-like by nature, which resolves the problems that occur if one
tries to implement bits back chaining with AC, which is queue-like. We now
describe this novel method, which we refer to as `Bits Back with ANS' (BB-ANS).

We can visualize the stack-like state of an ANS coder as

\newlength{\messageheight}
\setlength{\messageheight}{8pt}

\begin{tikzpicture}
  \draw (3, \messageheight) -- (0, \messageheight) -- (0, 0) -- (3, 0);
  \draw[dashed] (3, \messageheight) -- (3, 0);
\end{tikzpicture}

where the dashed line on the right symbolizes the encoding/decoding end or
`top' of the stack. When we encode a symbol $x$ onto the message stack we
effectively add it to the end, resulting in a `longer' state

\begin{tikzpicture}
  % Message
  \draw (3, \messageheight) -- (0, \messageheight) -- (0, 0) -- (3, 0) -- cycle;
  \draw (3, \messageheight) -- (5, \messageheight);
  \draw (3, 0)              -- (5, 0);
  \draw[dashed] (5, \messageheight) -- (5, 0);

  % Measure above message
  \draw (3, \messageheight + 10) -- (5, \messageheight + 10);
  \draw (3, \messageheight + 13) -- (3, \messageheight + 7 );
  \draw (5, \messageheight + 13) -- (5, \messageheight + 7 );
  \node [label=above:\small{\(\log1/P(x)\)}] at (4, \messageheight + 10) {};
\end{tikzpicture}

and when we decode (or equivalently, sample) a symbol \(x'\) from the stack we
remove it from the same end, resulting in a `shorter' state, plus the symbol
that we decoded.

\begin{tikzpicture}
  % Message
  \draw (1.5, \messageheight) -- (0, \messageheight) -- (0, 0) -- (1.5, 0);
  \draw [dashed] (1.5, \messageheight) -- (1.5, 0) -- (3, 0)
    -- (3, \messageheight) -- cycle;

  % Measure above message
  \draw (1.5, \messageheight + 10) -- (3, \messageheight + 10);
  \draw (1.5, \messageheight + 13) -- (1.5, \messageheight + 7 );
  \draw (3, \messageheight + 13) -- (3, \messageheight + 7 );
  \node [label=above:\small{\(\log1/P(x')\)}] at (2.25, \messageheight + 10) {};
\end{tikzpicture}\quad,\quad\(x'\)

For compactness, we use define a shorthand notation for encoding and decoding
operations: \(x\rightarrow P(\cdot)\) for encoding (pushing) \(x\) onto the stack using the distribution
\(P\), and \(x\leftarrow P(\cdot)\) for decoding (popping). The notation is
summarized in \Cref{tab:ans-notation}. \Cref{tab:sender_table} shows the states
of the message as the sender encodes a sample, using our bits back with ANS
algorithm, starting with an existing message, labelled `extra information', as
well as the sample $x$ to be encoded. The operations are performed starting at
the top of the table and working downwards.
\begin{table}[t]
\caption{The shorthand we use for ANS encoding and decoding operations, based
on left and right pointing arrows. The operations can be translated
mechanically to or from the pseudocode shown in the right-hand
column.}\label{tab:ans-notation}
\centering
  \small
  \begin{tabular}{lll}
    Notation                 &Meaning&Pseudocode\\\midrule
    \(x\rightarrow P(\cdot)\)&``Encode \(x\) using \(P\)''
                             &\texttt{message = push\_P(message, x)}\\
    \(x\leftarrow P(\cdot)\) &``Decode \(x\) using \(P\)''
                             &\texttt{message, x = pop\_P(message)}\\
  \end{tabular}
\end{table}
\begin{table}[h]
\centering
\caption{
Visualizing the process by which a sender encodes (pushes) a symbol $x$ onto an
ANS message stack using Bits Back with ANS. The process starts with an
existing ANS message, labelled `extra information', and the operations in the
right hand column are performed starting at the top of the table and working
downwards. The `Variables' column shows variables which are known to the sender
before each operation is performed.}
  \label{tab:sender_table}
    \small
  \begin{tabular}{lll}
%   \multicolumn{2}{c|}{Sender state} & \multicolumn{1}{c}{}\\
%   \cline{1-2}
  \multicolumn{1}{c}{BB-ANS message}&\multicolumn{1}{c}{Variables}&\multicolumn{1}{c}{Operation}\\
  \midrule&&\\
\begin{tikzpicture}
  \draw (3, 0) -- (0, 0) -- (0, \messageheight) -- (3, \messageheight);
  \draw [dashed] (3, 0) -- (3, \messageheight);
  \node [label=above:\small{Extra information}] at (1.5, \messageheight) {};
\end{tikzpicture}&$x$&\\&&
    \\
\begin{tikzpicture}
  \draw (1.2, 0) -- (0, 0) -- (0, \messageheight) -- (1.2, \messageheight);
  \draw [dashed] (1.2, 0) -- (3, 0) -- (3, \messageheight) -- (1.2,
    \messageheight) -- cycle;
  \draw (1.2, \messageheight + 10) -- (3, \messageheight + 10);
  \draw (1.2, \messageheight + 13) -- (1.2, \messageheight + 7 );
  \draw (3, \messageheight + 13) -- (3, \messageheight + 7 );
  \node [label=above:\small{\(\log1/Q(z\given x)\)}]
    at (2.1, \messageheight + 10) {};
\end{tikzpicture}&$x, z$&\(z\leftarrow Q(\cdot\given x)\)\\&&
    \\
    \begin{tikzpicture}
  \draw (2.5, 0) -- (0, 0) -- (0, \messageheight) -- (2.5, \messageheight);
  \draw (1.2, 0) -- (1.2, \messageheight);
  \draw [dashed] (2.5, 0) -- (2.5, \messageheight);
  \draw (1.2, \messageheight + 10) -- (2.5, \messageheight + 10);
  \draw (1.2, \messageheight + 13) -- (1.2, \messageheight + 7 );
  \draw (2.5, \messageheight + 13) -- (2.5, \messageheight + 7 );
  \node [label=above:\small{\(\log1/P(x\given z)\)}]
    at (1.85, \messageheight + 10) {};
  \end{tikzpicture}&$z$&\(x\rightarrow P(\cdot\given z)\)\\&&
    \\
    \begin{tikzpicture}
  \draw (4.1, 0) -- (0, 0) -- (0, \messageheight) -- (4.1, \messageheight);
  \draw (1.2, 0) -- (1.2, \messageheight);
  \draw (2.5, 0) -- (2.5, \messageheight);
  \draw [dashed] (4.1, 0) -- (4.1, \messageheight);
  \draw (2.5, \messageheight + 10) -- (4.1, \messageheight + 10);
  \draw (2.5, \messageheight + 13) -- (2.5, \messageheight + 7 );
  \draw (4.1, \messageheight + 13) -- (4.1, \messageheight + 7 );
  \node [label=above:\small{\(\log1/P(z)\)}]
    at (3.3, \messageheight + 10) {};
  \end{tikzpicture}&       &\(z\rightarrow P(\cdot)\)\\
  \end{tabular}
\end{table}

This process is clearly invertible, by reversing the order of operation and
replacing encodes with decodes and sampling with encoding. Furthermore it can
be repeated; the ANS message at the end of encoding is still an ANS message, and
therefore can be readily used as the extra information for encoding the next
symbol. The algorithm is compatible with any model whose prior, likelihood and
(approximate) posterior can be encoded and decoded with ANS, i.e. it is
necessary and sufficient to be able to compute conditional CDFs and inverse
CDFs under those distributions. A simple Python implementation of both the
encoder and decoder of BB-ANS is shown in \cref{bb-ans-code}.

\begin{figure}[h]
\begin{lstlisting}
def push(message, x):
    # (1) Sample $z$ according to $Q(z|x)$
    #       Decreases message length by $\log 1/Q(z|x)$
    message, z = posterior_pop(x)(message)

    # (2) Encode $x$ according to the likelihood $P(x|z)$
    #       Increases message length by $\log 1/P(x|z)$
    message = likelihood_append(z)(message, x)

    # (3) Encode $z$ according to the prior $P(z)$
    #       Increases message length by $\log 1/P(z)$
    message = prior_append(message, z)

    return message

def pop(message):
    # (3 inverse) Decode $z$ according to $P(z)$
    message, z = prior_pop(message)

    # (2 inverse) Decode $x$ according to $P(x|z)$
    message, x = likelihood_pop(z)(message)

    # (1 inverse) Encode $z$ according to $Q(z|x)$
    message = posterior_append(x)(message, z)

    return message, x
\end{lstlisting}
\caption{
  Python implementation of BB-ANS encode (`push') and decode (`pop') methods.}
\label{bb-ans-code}
\end{figure}

\section{Issues affecting the efficiency of BB-ANS}
A number of factors can affect the efficiency of compression with BB-ANS, and
mean that in practice, the coding rate will never be exactly equal to the ELBO.
For any algorithm based on AC/ANS, the fact that all probabilities have to be
approximated at finite precision has some detrimental effect. When encoding a
batch of only a small number of i.i.d.\ samples, with no `extra information' to
communicate, the inefficiency of encoding the first data point may be
significant. In the worst case, that of a batch with only one data point, the
message length will be equal to the log joint, $\log P(x, z)$. Note that
optimization of this is equivalent to maximum a posteriori (MAP) estimation.
However, for a batch containing multiple images, this effect is amortized.
\Cref{fig:visual_compression} shows an example with 30 samples, where BB-ANS
performs well.

Below we discuss two other issues which are specific to BB-ANS. We investigate
the magnitude of these effects experimentally in Section
\ref{mnist_experiments}. We find that when compressing the MNIST test set, they
do not significantly affect the compression rate, which is typically close to
the negative ELBO in our experiments.

\subsection{The extra information}\label{sec:plc-extra-info}
Extra information is required to initialize the bits-back chain. In practical
situations, there may not be any other information that we wish to
communicate, apart from that which we are modelling with a latent variable
model. In this situation, we can simply send random information at the start of
the chain. This means that the message has some redundancy, and the minimum
amount of redundant information which we must transmit scales with the
dimensionality of the latent variables. In the experiments which we present in
this chapter, we use small models with a latent dimension of 40 or 50, and find
that in this case the overhead is low enough that we can achieve good
compression even when compressing short sequences of 30 images (see
\cref{fig:visual_compression}). In \Cref{chap:hilloc}, we use far larger,
hierarchical models, with latent dimensionality in the hundreds of thousands,
and this becomes a serious issue. In \Cref{sec:starting} we describe a
straightforward method to deal with the overhead, by using a hybrid codec: we
first encode images using FLIF, which doesn't require extra bits, until enough
of a buffer has been built up, then when possible we switch to BB-ANS, which
achieves a better bit-rate.

\subsection{Discretizing a continuous latent space}\label{sec:plc-disc}
Bits back coding has previously been implemented only for models with discrete
latent variables, in \citet{frey1997}. However, many successful latent variable
models utilize continuous latents, including the VAE which we use in our
experiments. We present here a derivation, based on \citet{mackay2003}, of the
surprising fact that continuous latents can be coded with bits back, up to
arbitrary precision, without affecting the coding rate. We also briefly discuss
our implementation, which as far as we are aware is the first implementation of
bits back to support continuous latents. In the following we continue to use
upper case \(P\) and \(Q\) to denote mass functions for discrete distributions,
and use lower case \(p\) and \(q\) for density functions of continuous
distributions.

We can crudely approximate a continuous probability distribution, with density
function $p$, with a discrete distribution, by partitioning the real line into
`buckets' of equal width $\delta z$. Indexing the buckets with $i\in I$, we
assign a probability mass to each bucket of $P(i) \approx p(y_i)\delta z$,
where $y_i$ is some point in the $i^{\text{th}}$ bucket (say its centre).

During bits back coding, we must discretize both the prior and the approximate
posterior using the \emph{same set of buckets}. Sampling from the discrete
approximation $Q(i\given x)$ uses approximately $\log (q(y_i\given x)\delta z)$
bits, and then encoding according to the discrete approximation to the prior
$P$ costs approximately $\log (p(y_i)\delta z)$ bits. The expected message
length increase for bits back with a discretized latent is therefore
\begin{align}
  \Delta L
  = -\mathbb{E}_{Q(i\given x)} \bigg[ \log \frac{P(x\given
  y_i)p(y_i)\delta z}{q(y_i\given x)\delta z}\bigg].
\end{align} The $\delta
z$ terms cancel, and thus the only cost to discretization results from the
discrepancy between our approximation and the true, continuous, distribution.
However, if the density functions are smooth (as they are in a VAE), then for
small enough $\delta z$ the effect of discretization will be negligible.

Note that the number of bits required to generate the latent sample scales with
the precision $-\log\delta z$, meaning reasonably small precisions should be
preferred in practice. Furthermore, the benefit from increasing latent
precision past a certain point is negligible for most machine learning model
implementations, since they operate at 32 bit precision. In our experiments we
found that increases in performance were negligible past 16 bits per latent
dimension.

In our implementation, we divide the latent space into buckets which have equal
mass under the prior (as opposed to equal width). This discretization is simple
to implement, the computational complexity does not increase with the precision
of the number of discrete intervals, and it is efficient, for the following
reasons: firstly, because the discretized prior reduces to a discrete uniform
distribution, which is cheap to encode/decode; secondly, because the posterior
we use is parametric (a Gaussian), we can do all the computation necessary for
encoding and decoding without enumerating all intervals; and thirdly (and more
subtlely), we don't have to worry about encoding discrete intervals which have
zero mass.  This is because the prior is uniform (so all intervals have the
same, nonzero mass), and the only intervals we ever need to encode with the
discretized posterior have been \emph{sampled} from the discretized posterior,
and thus cannot have zero mass.  This saves us from a costly process of
ensuring that no symbols have zero mass. Note that this discretization method
can only be applied when dimensions of a latent vector are independent random
variables. In \Cref{sec:discretization} we show how to extend this simple
discretization to latents with non-trivial dependence structure.

\subsection{The need for `clean' bits}\label{clean_bits}
In our description of bits back coding in Section \ref{bbc}, we noted that the
`extra information' required to seed bits back should take the form of `random
bits'. More precisely, we need the result of mapping these bits through our
decoder to produce a true sample from the distribution $q(z\given x)$. A
sufficient condition for this is that the bits are i.i.d.\ Bernoulli
distributed with probability $\frac{1}{2}$ of being in each of the states $0$
and $1$\footnote{This is sufficient because of a fundamental
property of ANS; that integers which decode to a symbol \(x\) are distributed
within the natural numbers with density \(P(x)\).}. We refer to such bits as
`clean'.

During chaining, we effectively use each compressed data point as the seed for
the next. Specifically, we use the bits at the top of the ANS stack, which are
the result of coding the previous latent $z$ according to the prior $p(z)$.
Will these bits be clean? The latent $z$ is originally generated as a sample
from $q(z\given x)$. This distribution is clearly not equal to the prior,
except in degenerate cases, so naively we wouldn't expect encoding $z$
according to the prior to produce clean bits. However, the true sampling
distribution of $z$ is in fact the \emph{average} of $q(z\given x)$ over the
data distribution. That is, $q(z)\triangleq\sum_x q(z\given
x)P_\mathrm{data}(x)$. This is referred to in \citet{hoffman2016} as the
`average encoding distribution'.

If $q$ is equal to the true posterior, and the model is perfectly fit to the
data, then $q(z)\equiv p(z)$, however in general neither of these conditions
are met.  \citet{hoffman2016} measure the discrepancy empirically using what
they call the `marginal KL divergence' $\kl{q(z)}{p(z)}$, showing that this
quantity contributes significantly to the ELBO for three different VAE like
models learned on MNIST.  This difference implies that the bits at the top the
ANS stack after encoding a sample with BB-ANS will not be perfectly clean,
which could adversely impact the coding rate. However, empirically we have not
found this effect to be significant.

\section{Experiments}\label{sec:plc-experiments}
\subsection{Using a VAE as the latent variable model}
We now demonstrate the BB-ANS coding scheme using a VAE. This model has a
multidimensional latent with standard Gaussian prior and diagonal Gaussian
approximate posterior:
\begin{align}
  p(z)         &= N(z; 0, I)\\
  q(z\given x) &= N(z; \mu(x), \text{diag}(\sigma^2(x))).
\end{align}
We choose an output distribution (likelihood) $P(x\given z)$ suited to the
domain of the data we are modelling (see below). The usual VAE training
objective is the ELBO, which, as we noted in \Cref{sec:bbc-sub}, is the
negative of the expected message length with bits back coding. We can therefore
train a VAE as usual and plug it into the BB-ANS framework without modification.

\subsection{Compressing MNIST}\label{mnist_experiments}
We consider the task of compressing the MNIST dataset \citep{lecun1998}. We
first train a VAE on the training set and then compress the test set using
BB-ANS with the trained VAE\@. The MNIST dataset has pixel values in the range
of integers 0, \ldots, 255. As well as compressing the raw MNIST data, we also
present results for stochastically binarized MNIST \citep{salakhutdinov2008}.
For both tasks we use VAEs with fully connected generative and recognition
networks, and ReLU activations.\looseness=-1

For binarized MNIST the generative and recognition networks each have a single
deterministic hidden layer of dimension 100, with a stochastic latent of
dimension 40. The generative network outputs logits parametrizing a Bernoulli
distribution on each pixel. For the full (non-binarized) MNIST dataset each
network has one deterministic hidden layer of dimension 200 with a stochastic
latent of dimension 50. The output distributions on pixels are modelled by a
beta-binomial distribution, which is a two parameter discrete distribution. The
generative network outputs the two beta-binomial parameters for each pixel.

We initialize the BB-ANS chain with a supply of `clean' bits. We find that
around 400 bits are required for this in our experiments. The precise number
of bits required to start the chain depends on the entropy of the discretized
approximate posterior (from which we are initially sampling), and scales
roughly linearly with the dimensionality of the latents.

We report the achieved compression against a number of benchmarks in Table
\ref{tab:results}. Despite the relatively small network sizes and simple
architectures we have used, the BB-ANS scheme outperforms benchmark compression
schemes. While it is encouraging that even a relatively small latent variable
model can outperform standard compression techniques when used with BB-ANS, the
more important observation to make from Table \ref{tab:results} is that the
achieved compression rate is very close to the value of the negative test ELBO
seen at the end of VAE training.

\begin{table}[t]
    \centering
    \small
    \begin{tabular}{llcc}
                    &    &Binarized MNIST&Full MNIST\\\midrule[\heavyrulewidth]
                    &Raw &1              &8         \\\midrule
      \emph{Generic}&bz2 &0.25           &1.42      \\
                    &gzip&0.33           &1.64      \\\midrule
      \emph{Image}  &PNG &0.78           &2.79      \\
                    &WebP&0.44           &2.10      \\\midrule
      \emph{Models} &Locally masked PixelCNN\tablefootnote{
                     \citet{jain2020}.}&(0.14)&(0.65)\\
                                       &Our small VAE&(0.19)&(1.39)\\\midrule
                     \emph{BB-ANS} &Our small VAE&0.19&1.41
    \end{tabular}
    \caption{
    Compression rates on the binarized MNIST and full MNIST test sets, using
    BB-ANS and other benchmark compression schemes, measured in bits per
    dimension. Note that PNG and WebP are included for context but the
    comparison is not particularly fair, because the image files contain
    metadata, the size of which is significant relative to the size of an MNIST
    image, particularly in the binarized case. We also give (in brackets) the
    negative ELBO value for our trained VAEs, and the log-likelihood under the
    current state of the art model, `Locally masked PixelCNN', all evaluated on
    the test set.
    }
    \label{tab:results}
\end{table}

\begin{figure}[h]
\centering
\begin{subfigure}{.5\textwidth}
  \centering
  \includegraphics[width=\linewidth]{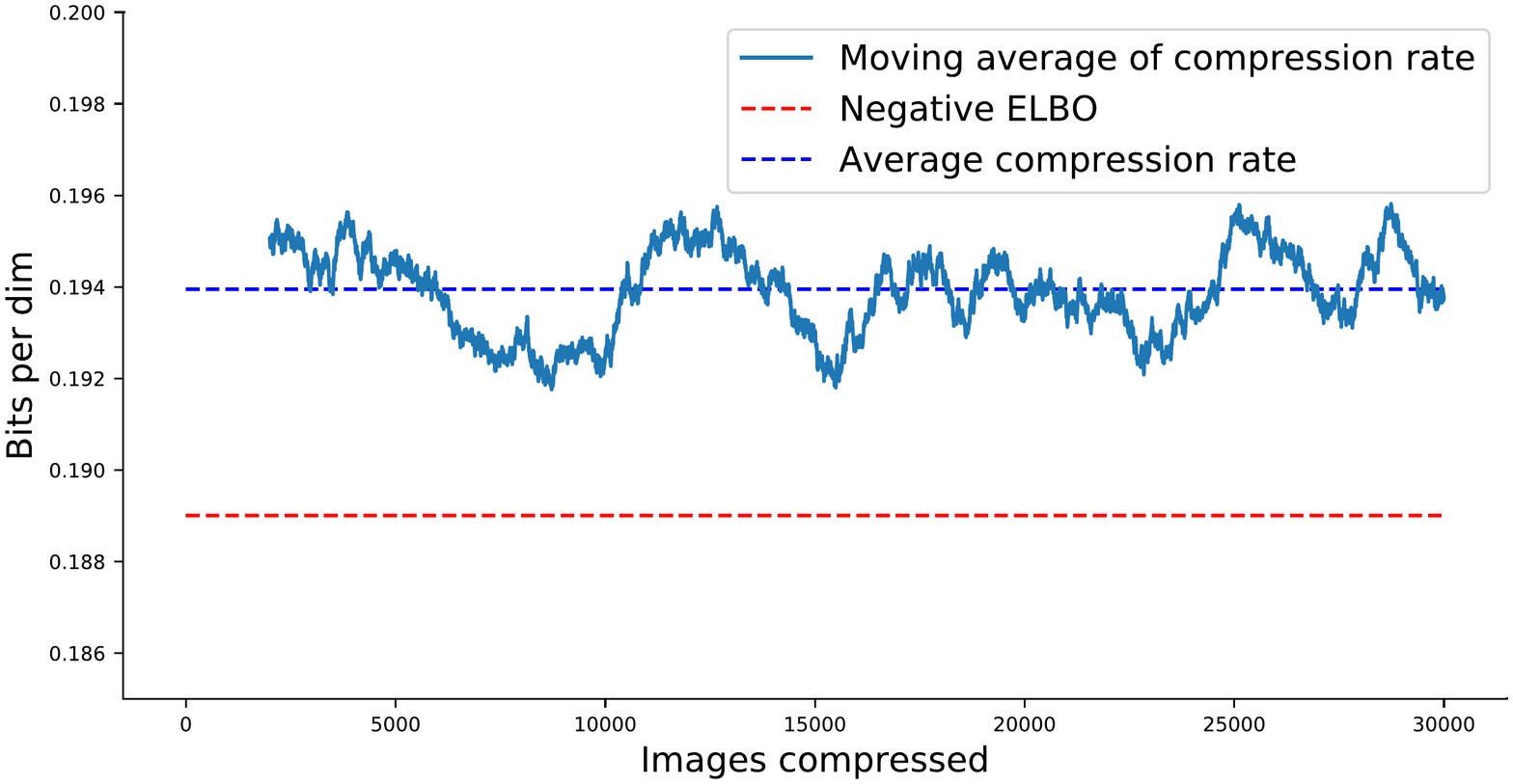}
  \caption{Binarized MNIST}
  \label{fig:sub1}
\end{subfigure}%
\begin{subfigure}{.5\textwidth}
  \centering
  \includegraphics[width=\linewidth]{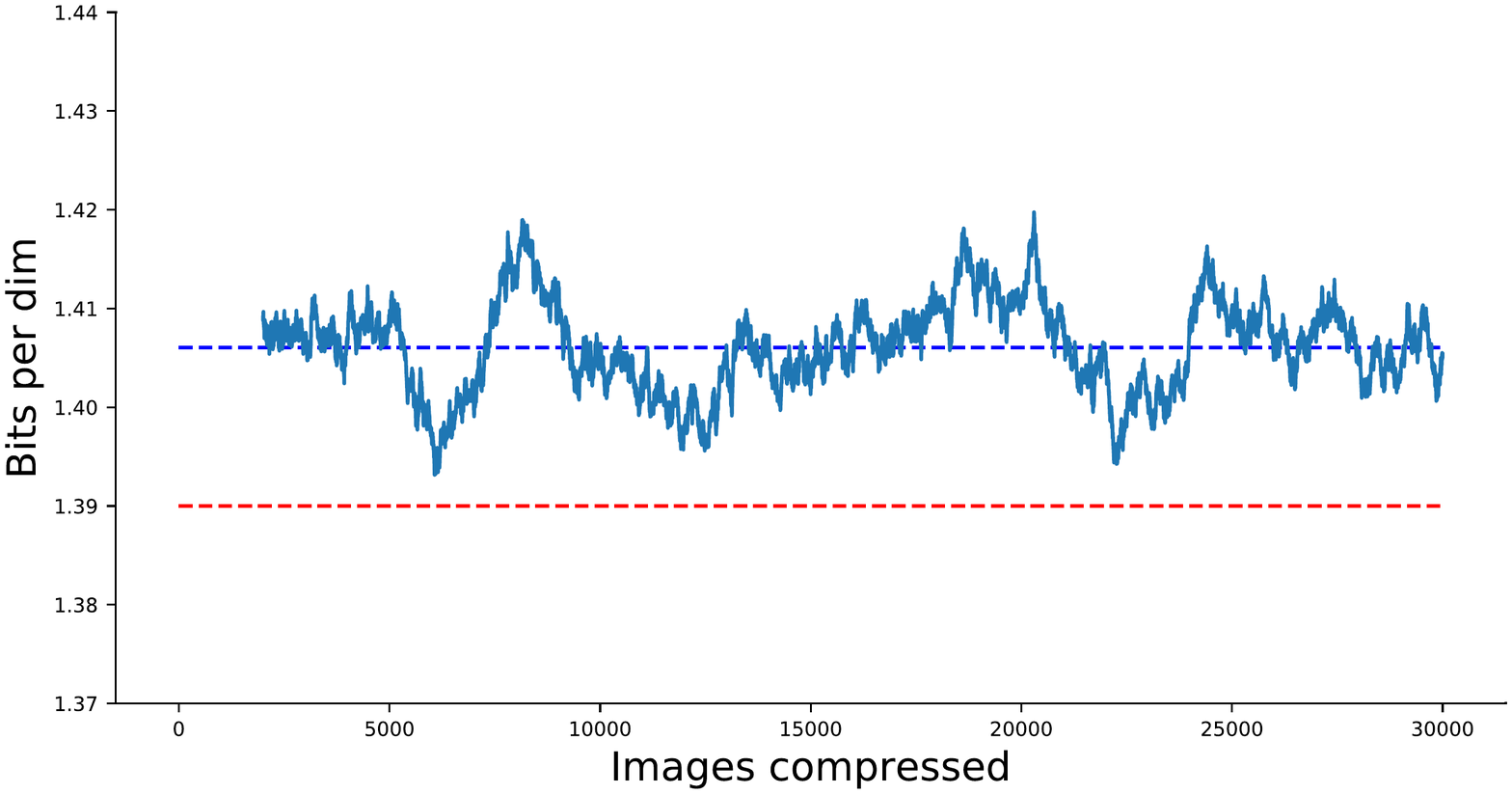}
  \caption{Full MNIST}
  \label{fig:sub2}
\end{subfigure}
\caption{
  A 2000 point moving average of the compression rate, in bits per dimension,
  during the compression process using BB-ANS with a VAE. We compress a
  concatenation of three shuffled copies of the MNIST test set.
}
\label{fig:dirty_bits}
\end{figure}

In particular, the detrimental effects of finite precision, the extra
information overhead (\cref{sec:plc-extra-info}), discretizing the latent
(Section \ref{sec:plc-disc}) and of less `clean' bits (Section \ref{clean_bits}) do
not appear to be significant. Their effects can be seen in Figure
\ref{fig:dirty_bits}, accounting for the small discrepancy of around $1\%$
between the negative ELBO and the achieved compression.

\section{Discussion}
\subsection{Extending BB-ANS to state-of-the-art latent variable models}
Implementing a state-of-the-art latent variable model is not the focus of this
chapter. However, we have shown that BB-ANS can compress data to sizes very
close to the negative ELBO for small-scale models. In \Cref{chap:hilloc} we
demonstrate BB-ANS on a large-scale model and achieve state-of-the-art lossless
compression on images from the ImageNet dataset.

Another extension of BB-ANS is to latent Gaussian state space models such as
those studied in \citet{johnson2016}, and to state space models more generally.
Very recent work (which originated from discussions during the viva for this
PhD) has shown how to do this by interleaving push/pop steps with the
time-steps in a model \citep{townsend2021}.

\subsection{Parallelization of BB-ANS}
Modern machine learning models are optimized to exploit batch-parallelism and
model-parallelism and run fastest on GPU hardware. Almost all of the
computation in the BB-ANS algorithm could be executed in parallel, on GPU
hardware, and the arithmetic operations required for ANS coding should not be a
computational bottleneck in a system which uses a deep generative model with
ANS. In \Cref{chap:vectorizing} we give more detail on how to implement ANS on
parallel hardware, and present our own fast CPU implementation, written in
NumPy \citep{oliphant2015}, based on ideas from \citet{giesen2014}.

\subsection{Communicating the model}
A neural net based model such as a VAE may have many thousands of parameters.
Although it is not the focus of this work, the cost of communicating and
storing a model's parameters would need to be considered when developing a
system which uses BB-ANS with a large scale model.

However, we can amortize the one-time cost of communicating the parameters over
the size of the data we wish to compress. If a latent variable model could be
trained such that it could model a wide class of images well, then BB-ANS could
be used in conjunction with such a model to compress a large number of images.
This makes the cost of communicating the model weights worthwhile to reap the
subsequent gains in compression. Efforts to train latent variable models to be
able to model such a wide range of images are currently of significant interest
to the machine learning community, for example on expansive datasets such as
ImageNet \citep{deng2009}. We therefore anticipate that this is the most
fruitful direction for practical applications of BB-ANS.

We also note that there have been many recent developments in methods to
decrease the space required for neural network weights, without hampering
performance. For example, methods involving quantizing the weights to low
precision \citep{han2016,ullrich2017}, sometimes even down to single
bit precision \citep{hubara2016}, are promising avenues of research that could
significantly reduce the cost of communicating and storing model weights.

\section{Conclusion}
Probabilistic modelling of data is a highly active research area within machine
learning. Given the progress within this area, it is of interest to study the
application of probabilistic models to lossless compression. Indeed, if
practical lossless compression schemes using these models can be developed then
there is the possibility of significant improvement in compression rates over
existing methods.

In this chapter we have shown the existence of a scheme, BB-ANS, which can be
used for lossless compression using latent variable models. We demonstrated
BB-ANS by compressing the MNIST dataset, achieving compression rates superior
to generic algorithms.  We have shown how to handle the issue of latent
discretization. Crucially, we were able to compress to sizes very close to the
negative ELBO for a large dataset. This is the first time this has been
achieved with a latent variable model, and suggests that state-of-the-art latent
variable models could be used in conjunction with BB-ANS to achieve
significantly better lossless compression rates than current methods. All
components of BB-ANS are readily parallelizable, and in
\Cref{chap:vectorizing}, we describe a method for vectorizing ANS, before
scaling BB-ANS up to large models and full-size colour photographs in
\Cref{chap:hilloc}.

%% file: Vectorizing.tex
\chapter{Vectorizing ANS with Craystack}\label{chap:vectorizing}
To implement the prototype compression system presented in \Cref{chap:hilloc},
we wrote new software for performing vectorized ANS operations, using NumPy
\citep{oliphant2015}. We call the software tool which we wrote `Craystack', and
make it available open source at \jurl{github.com/j-towns/craystack} (mirrored
at \jurl{doi.org/10.5281/zenodo.4707276}). Tom Bird and Julius Kunze both
assisted with the implementation of Craystack; Tom also created
\cref{fig:craystack-diagram}. This chapter gives low-level detail on how ANS
vectorization works, and a high-level overview of Craystack, as well as
discussion of future directions for compression prototyping software.

We begin by describing how ANS vectorization works. The method is based on
\citet{giesen2014}. You will need to have read \Cref{chap:ans} to understand
\Cref{sec:vectorizing-ans}.

\section{Vectorizing asymmetric numeral systems}\label{sec:vectorizing-ans}
We now describe a method for `vectorizing' ANS, that is, generalizing the
\push\ and \pop\ operations and expressing them in terms of vector or
`single-instruction-multiple-data' (SIMD) functions. What we give here is a
high-level description, with pseudo-code. For a full implementation, which uses
NumPy, see the code repository linked above.

\subsection{The vectorized message data structure}
\begin{figure}[ht]
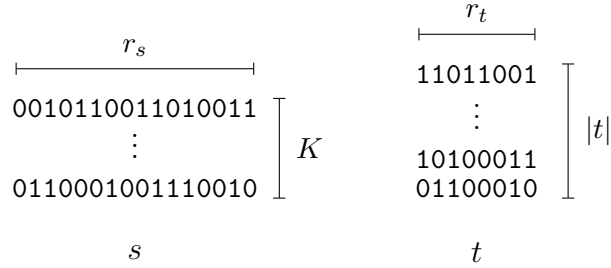

  \centering
  \drawvectormessage
  \caption{
    The two components of a (vectorized) message: the vector of unsigned
    integers \(s\) (with \(r_s = 16\)) and the stack of unsigned integers \(t\)
    (with \(r_t = 8\)). The integers are represented here in base 2 (binary).
}
  \label{fig:vector-message}
\end{figure}

The data structure for the message in vectorized ANS is identical to the data
structure described in \Cref{sec:message}, except that instead of a scalar
\(s\) we use a vector \(s = (s_1, s_2, \ldots, s_K)\). We refer to \(K\) as the
`size' of the message (not to be confused with the message \emph{length}). A
diagram of a vectorized message is shown in \cref{fig:vector-message}. In our
implementation we use a NumPy array to represent \(s\).

\subsection{Vectorizing the pop operation}
The original (scalar) version of the \pop\ function was defined in
\Cref{sec:scalar-pop}, as
\begin{singlespace}
\begin{lstlisting}[frame=single]
def pop($m$):
    $s$, $t$ := $m$
    $s'$, $x$ := $d$($s$)
    $s$, $t$ := renorm($s'$, $t$)
    return ($s$, $t$), $x$
\end{lstlisting}
\end{singlespace}
we needn't alter this high level definition, but just need to look at the
functions \(d\) and \texttt{renorm}. The original definition of \(d\) is
\begin{singlespace}
\begin{lstlisting}[frame=single]
def $d$($s$):
    $\bar s$ := $s\bmod 2^{r}$
    $x$, $c$, $p$ := $f_X(\bar s)$
    $s'$ := $p \cdot (s \div 2^{r}) + \bar s - c$
    return $s'$, $x$
\end{lstlisting}
\end{singlespace}
This is actually trivial to vectorize, simply by replacing all scalar
arithmetic operations by their (element-wise) vector counterparts. We do not
need to change the definition in any way, as long as we use arithmetic
operators which are compatible with vector arguments. Note that our vectorized
\(d\) function applied in this way will produce a vector \(x = (x_1, \ldots,
x_K)\), and the resulting \(s'\) will also be a vector of length \(K\).

The scalar function \texttt{renorm} was defined in \Cref{sec:scalar-pop} as
\begin{singlespace}
\begin{lstlisting}[frame=single]
def renorm($s$, $t$):
    while $s < 2^{r_s - r_t}$:
        $t$, $t_{\mathrm{top}}$ := stack_pop($t$)
        $s$ := $2^{r_t} \cdot s$ + $t_{\mathrm{top}}$
    return $s$, $t$
\end{lstlisting}
\end{singlespace}
For vector \(s\) this can be generalized to
\begin{singlespace}
\begin{lstlisting}[frame=single]
def renorm($s$, $t$):
    for $s_k$ in $s$:
        while $s_k < 2^{r_s - r_t}$:
            $t$, $t_{\mathrm{top}}$ := stack_pop($t$)
            $s_k$ := $2^{r_t} \cdot s_k$ + $t_{\mathrm{top}}$
    return $s$, $t$
\end{lstlisting}
\end{singlespace}
Implementing this function and its inverse, in NumPy, is difficult, so we make
the simplifying assumption that \(r_s \leq 2r_t\), where \(r_s\) and \(r_t\)
are the precisions of the unsigned integers \(s\) and \(t\), respectively; as
in \Cref{chap:ans}. This assumption has only a negligible effect on the
compression rate at the precisions we typically use (\(r_s=64\) and
\(r_t=32\)), and implies that at most one iteration of the inner while loop
will ever be required for each element, so the definition of \texttt{renorm}
becomes
\begin{singlespace}
\begin{lstlisting}[frame=single]
def renorm($s$, $t$):
    for $s_k$ in $s$:
        if $s_k < 2^{r_s - r_t}$:
            $t$, $t_{\mathrm{top}}$ := stack_pop($t$)
            $s_k$ := $2^{r_t} \cdot s_k$ + $t_{\mathrm{top}}$
    return $s$, $t$
\end{lstlisting}
\end{singlespace}
replacing \texttt{while} with \texttt{if}. That version of \texttt{renorm} is
straightforward to implement using the NumPy \texttt{where} function and
boolean indexing.

The derivation of a vectorized \push\ function by inverting \pop\ is
mechanical, and rather than detailing it here we refer the interested reader to
the implementation in the file \texttt{craystack/rans.py} in the Craystack
repository.

\subsection{The length of the vectorized message}
In \Cref{sec:message}, we defined the \emph{length} of a scalar
message as
\begin{equation}
  l(m) := r_s + r_t\abs{t}
\end{equation}
and the \emph{effective length} as
\begin{equation}
  l^*(m) := \log s + r_t\abs{t}.
\end{equation}
The length of the scalar message tells us how many bits are required to store a
flattened representation of that message. We could flatten a vector message \(m
= ((s_1, \ldots, s_K), t)\), simply by concatenating the elements
\(s_1,\ldots,s_K\) and the elements of \(t\), which would lead to a flattened
string with length
\begin{equation}
  l_\text{naïve}(m) = Kr_s + r_t\abs{t}.
\end{equation}
We can improve on this length, by approaching the question of how to flatten
\(m\) as a \emph{lossless compression problem}. We split the vector \(s\) into
a pair \(s_1, (s_2,\ldots,s_K)\), and treat \(m' = (s_1, t)\) as a
\emph{scalar} message, and the question becomes, how best to \emph{push} the
elements \(s_2,\ldots,s_K\) onto \(m'\)? We can use ANS (a near-optimal
compressor) to do this, and then we just have to pick a distribution to use to
model \(s_2,\ldots,s_K\).

\subsubsection{The Benford distribution}
It has been observed empirically, by \citet{bloom2014}, that in practical
settings, the elements \(s_k\) precisely follow \emph{Benford's law}
\citep{benford1938}.  Benford's law says that samples of leading digits from
sets of `real-world' numbers tends to follow a distribution with mass function
\begin{equation}
  P_\text{Benford}(x)\propto 1 / x,
\end{equation}
where \(x\) is the integer formed by concatenating the \(r\) leading digits of
a sampled constant, for some fixed value of \(r\) (the normalizing constant of
\(P\) is a function of \(r\)). We refer to this as the `Benford distribution'.
\citet{mackay2003} argues that the reason that digits in physical constants
must follow this distribution is that their distribution ought not to depend on
the units of measurement used, and that the Benford distribution is the unique
distribution which is invariant to rescaling of units.

The number \(s\) can be thought of as the first \(r_s / (r_s - r_t)\) digits of
a flat message, if the message is treated as a large integer, expressed in base
\(r_s - r_t\). Since, as we have shown, message length is approximately equal
to the information content of an encoded sequence plus a constant, the fact
that \(s\) is Benford distributed is equivalent to saying that
\emph{probability masses of sample sequences tend to follow Benford's law}.

Why might this be? Similarly to \citet{mackay2003}, we can sketch a proof by
contradiction. Consider the sequence
\begin{equation}
 p_n = \prod_{i=1}^n P(x_i\given x_1,\ldots,x_{i-1})
\end{equation}
for a sample \(x_1, x_2, \ldots, x_n, \ldots\) with mass function \(P\).  Let
\(q_n\) be the integer composed of the first \(r\) non-zero digits of \(p_n\).
If we assume that \(q_n\) tends to a stationary distribution as \(n\)
increases, and that the stationary distribution is positive (that is, each
possible state of \(q_n\) has non-zero probability), then by definition the
stationary distribution must be invariant to the scale of \(p_n\), because for
each \(n\), \(q_{n+1}\) is derived from a rescaling of \(p_n\) (specifically,
multiplying \(p_n\) by \(P(x_{n+1}\given x_1,\ldots,x_n)\)).

We can approximate the normalizer for the Benford distribution over \(s\)
\begin{align}
  \sum_{s=2^{r_s - r_t}}^{2^{r_s}} \frac{1}{s}
    &\approx \int_{2^{r_s - r_t}}^{2^{r_s}}\frac{1}{s}\\
    &= [r_s - (r_s - r_t)]\ln2\\
    &= r_t\ln2
\end{align}
and thus, using the Benford distribution to append the elements
\(s_2,\ldots,s_K\) to the scalar message \((s_1, t)\) results in a flattened
message whose length satisfies
\begin{equation}\label{eq:vec-opt-length}
  l_\mathrm{opt}(m) \leq r_s + \sum_{k=2}^{K} \log s_k + (K-1)\log (r_t\ln2) +
  r_t\abs{t} + (K-1)\epsilon,
\end{equation}
by applying \cref{eq:pop-inequality} to each \(s_2, \ldots, s_K\).

\subsection{The cost of vectorization}
In the scalar case we were able to prove the following bound on message length,
\cref{eq:length-bound},
\begin{equation}\label{eq:scalar-opt-length}
    l(m) \leq h(x_1, \ldots, x_N) + N\epsilon + r_s.
  \end{equation}
If we initialize a vector message to \(m = (s_\mathrm{init}, t_\mathrm{init})\)
where \(s_\mathrm{init} := (2^{r_s - r_t}, 2^{r_s - r_t}, \ldots, 2^{r_s -
r_t})\), i.e. the minimum permissible value for each element, then apply
\cref{eq:vec-opt-length} and \cref{eq:pop-inequality} to each encoding
iteration element-wise, we get an inequality which is analogous to
\cref{eq:scalar-opt-length}
\begin{equation}
  l_\mathrm{opt}(m) \leq h(x_1, \ldots, x_N) + (K-1)\log (r_t\ln2) +
  K(N\epsilon + r_s - r_t) + r_t.
\end{equation}
For short messages and large \(K\), the overhead terms \((K-1)\log(r_t\ln2)\)
and \(K(r_s - r_t)\) may be significant.

One approach to reducing these overheads, which we use in \Cref{chap:hilloc},
is to encode the first samples in a sequence using \emph{scalar} ANS, then
expand the size of the message progressively by \emph{decoding} (i.e. sampling)
new elements \(s_k\) from the message as it grows in length. The \(K(r_s-r_t)\)
overhead term from the vector initial message disappears when we do this, and
if we use the Benford distribution to sample the \(s_k\) then the \((K-1)
\log(r_t\ln2)\) term is cancelled out as well, meaning we can vectorize with
negligible overhead in compression rate. However, this approach is complicated
to implement, and it is unclear whether it would be sensible to use it in
practical settings given the additional complexity required.\looseness=-1

\subsection{The benefit of vectorization}
\begin{figure}[ht]
\centering
\includegraphics[width=0.55\textwidth]{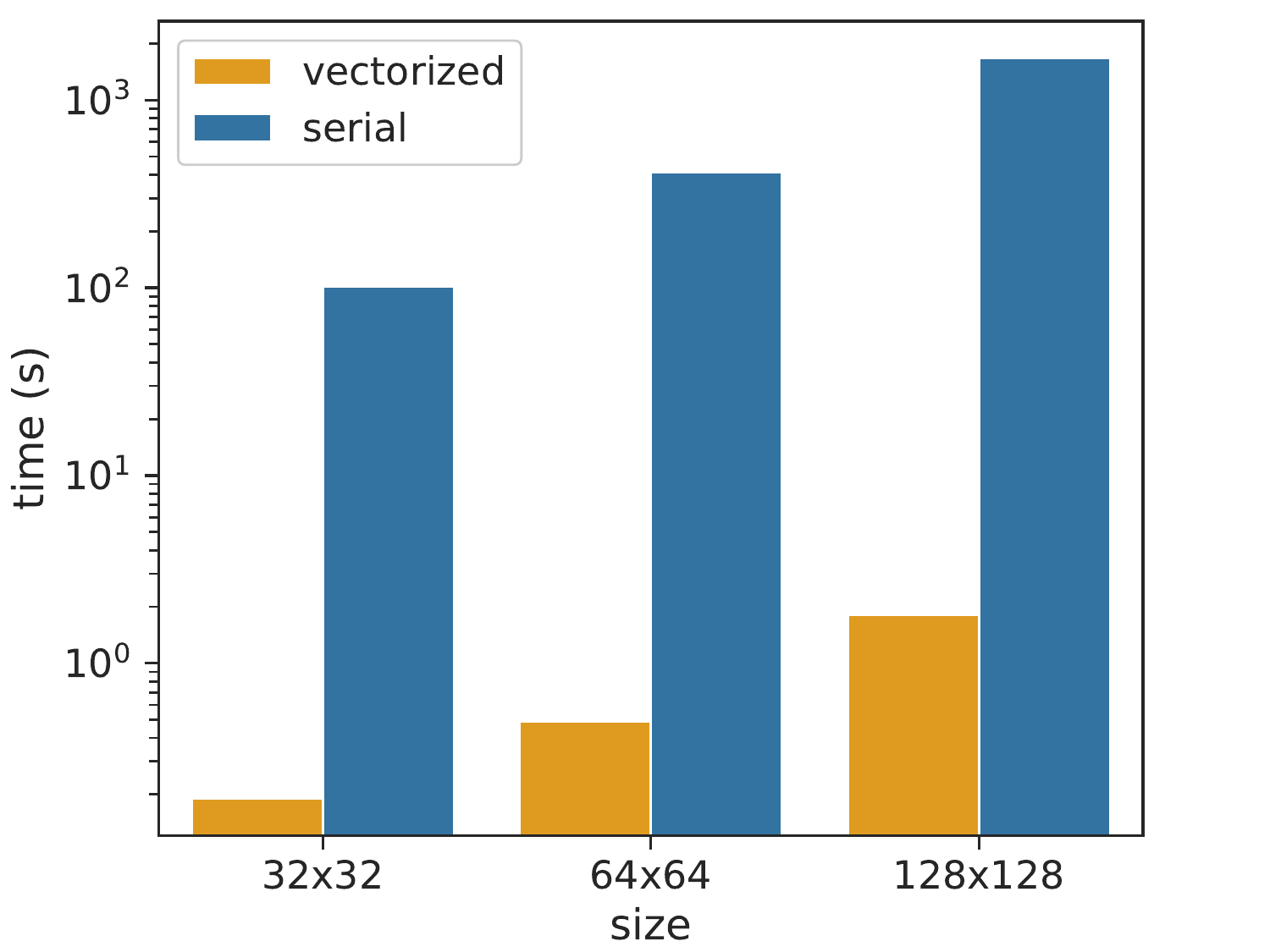}
\caption{
Runtime of vectorized vs.\ serial ANS implementations for different image
sizes. Times were computed on a desktop with 6 CPU cores.}
\label{fig:timings}
\end{figure}

The performance benefits of vectorization are drastic. To demonstrate this we
benchmarked the ANS encoding and decoding of ImageNet images (see
\Cref{chap:hilloc} for more details on the codec), using a scalar ANS
implementation vs a vectorized implementation. The results are shown in
\cref{fig:timings}, and show that in this setting the vectorized ANS
implementation was nearly three orders of magnitude faster than the scalar
implementation. This is because of loops which occur in the Python interpreter
in the scalar version being `pushed down' into highly optimized NumPy kernels,
implemented in C and Fortran.

\section{Craystack}
\begin{figure}[t]
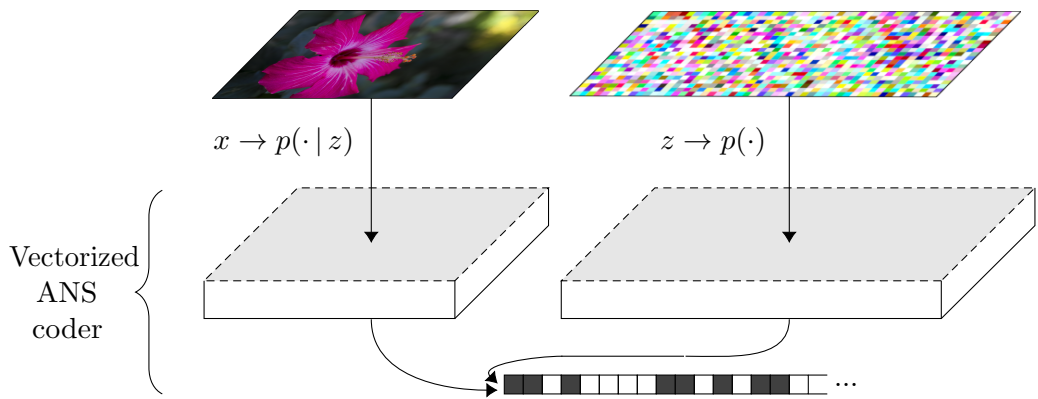

  \makebox[\textwidth][c]{\drawcraystackdiagram}
\caption{
Visualizing the process of pushing images and latents from a VAE to the
vectorized ANS stack with Craystack. The ANS stack head is shaped such that
images and latents can be pushed and popped in parallel, without reshaping.
Beneath the shaped top of the stack is the flat message stream output by ANS.}
\label{fig:craystack-diagram}
\end{figure}
Having described some of the low-level details of vectorized ANS, we now
discuss some of the high-level features of the Craystack library, which aims to
make prototyping lossless compression systems with ANS straightforward and
approachable for machine learning practitioners.
\subsection{High-level design}
When writing a lossless compression system, it is essential to ensure that the
decoder function is the precise inverse of the encoder. Manually writing code
which satisfies this constraint requires care, and one of the central goals of
Craystack is to ease this burden by ensuring that this constraint is satisfied
\emph{automatically}. To achieve this, Craystack provides a set of primitive
codecs, including codecs for coding data according to uniform, Bernoulli,
discretized logistic, discretized Gaussian, and categorical distributions. Each
primitive codec is comprised of a \((\push, \pop)\) inverse pair. As well as
the primitives, we also provide a set of combinators for composing and nesting
codecs. Examples of high level combinators included in Craystack are a BB-ANS
combinator and a combinator for compression with an auto-regressive model.

The code for the BB-ANS combinator is shown in \cref{fig:bb-ans-combinator}. By
using the Craystack primitives combined with the provided combinators, users
are able to build elaborate compressors without needing to worry about the
correctness of their code.

\begin{figure}[h]
\begin{lstlisting}[frame=single]
def BBANS(prior, likelihood, posterior):
    def push(message, data):
        message, latent = posterior(data).pop(message)
        message = likelihood(latent).push(message, data)
        message = prior.push(message, latent)
        return message

    def pop(message):
        message, latent = prior.pop(message)
        message, data = likelihood(latent).pop(message)
        message = posterior(data).push(message, latent)
        return message, data
    return Codec(push, pop)
\end{lstlisting}
\caption{
The BB-ANS combinator provided in Craystack. It accepts codecs for the prior,
likelihood and approximate posterior as arguments and returns a codec which
uses BB-ANS to compress data modelled with a latent variable model.
}\label{fig:bb-ans-combinator}
\end{figure}

\subsection{Generalizing vectorized ANS}
Another way in which Craystack aims for convenience is by allowing arbitrarily
shaped data arrays to be coded directly. The `vector' \(s\) in the Craystack
ANS implementation is a 1D NumPy array, but we allow codecs to use arbitrary
shaped NumPy views of \(s\), and multiple views of \(s\) can even be separated
into an arbitrary nesting of Python tuples, lists and dictionaries.
\Cref{fig:craystack-diagram} shows a visualization of this, where \(s\) is
split into a pair of rectangular views.

The API for this mechanism, which is via a function called \texttt{substack},
only requires the user to supply a function which creates the view from a flat
vector \(s\). For example, codecs for the latent and observation in
\cref{fig:craystack-diagram} could be expressed as
\begin{singlespace}
\begin{lstlisting}
latent_view = lambda s: s[:latent_size].reshape(latent_shape)
obs_view    = lambda s: s[latent_size:].reshape(obs_shape)

flat_latent_codec = substack(shaped_latent_codec, latent_view)
flat_obs_codec    = substack(shaped_obs_codec,    obs_view)
\end{lstlisting}
\end{singlespace}
The view functions are linear, orthonormal mappings (in the sense of linear
algebra), and therefore they can be \(inverted\) using reverse mode automatic
differentiation (AD). This is because the derivative of a linear function is
the function itself; reverse mode AD computes the transpose of the derivative,
and the transpose of an orthonormal mapping is equal to the mapping's (pseudo)
inverse. The \texttt{substack} function automatically computes the inverse of
the view functions using Autograd \citep{maclaurin2015}, ensuring that the
\push\ and \pop\ functions produced are inverse to one-another.  Methods for
implementing automatic view inversion from scratch (i.e.\ without exploiting
an AD system), can be found in \citet{voigtlander2009}.

\section{Future directions for Craystack}
We broadly see two major directions for improving Craystack. The first is to
allow ANS to run on accelerators such as GPUs. One way to achieve this would be
to port the implementation into Google's JAX software \citep{bradbury2018},
which enables use of GPU and TPU backends with a NumPy-like API\@. JAX also
allows just-in-time compilation with automatic loop fusion, amongst other
optimizations, and implementing deep generative models in JAX is reasonably
straightforward\footnote{During the very recent work described in
        \citet{ruan2021}, we implemented a proof-of-concept version of
        Craystack's core vectorized rANS in JAX, and have run it on GPU, though
        we haven't thoroughly benchmarked it yet. The implementation is
        available at \jurl{github.com/j-towns/crayjax}, mirrored at
\jurl{https://doi.org/10.5281/zenodo.4650348}.}.

One interesting open question is whether it is possible to have a user write a
decoder (\pop) function (which would be very similar to a function for sampling
from a model), and to automatically transform that function into an encoder
(\push) function.  Internally, JAX is based on a core system for writing
composable function transformations, with transformations such as forward and
reverse mode differentiation, auto-vectorization and just-in-time compilation
provided. More investigation is needed to find out whether this transformation
can be implemented in a practical manner, but if so it could fit well into the
existing JAX system.

%% file: HiLLoC.tex
\chapter{Scaling up bits back coding with asymmetric numeral systems}
\label{chap:hilloc}
This chapter is based on the paper `HiLLoC: lossless image compression with
hierarchical latent variable models', published at ICLR 2020
\citep{townsend2020}. We show that the bits back with ANS method presented in
\Cref{chap:plc} can be scaled up to larger models and applied to compression of
colour photographs, achieving a state-of-the-art compression rate on full-sized
images from the ImageNet dataset \citep{russakovsky2015}. This was a
collaboration with Tom Bird, who contributed to the ideas and wrote the paper
and experiments with me, and Julius Kunze, who mostly assisted in writing the
experiments.

In order to scale up the methods in \Cref{chap:plc}, we will use four novel
ideas:
\begin{enumerate}
    \item A vectorized ANS implementation supporting dynamic
      shape.\label{item:vec}
    \item Direct coding of arbitrary sized images using a fully convolutional
      model.\label{item:conv}
    \item Dynamic discretization of hierarchical latents.\label{item:dynamic}
    \item Initializing the bits back chain using a different
      codec.\label{item:init}
\end{enumerate}

We have already discussed \cref{item:vec} in detail in \Cref{chap:vectorizing},
and will discuss \cref{item:conv,item:dynamic,item:init} in
\Cref{sec:contributions}. We call the combination of BB-ANS using a
hierarchical latent variable model and the above techniques: `Hierarchical
Latent Lossless Compression' (HiLLoC). In our experiments (Section
\ref{sec:hilloc-experiments}), we demonstrate that HiLLoC can be used to
compress colour images from the ImageNet test set at rates close to the ELBO,
outperforming all of the other codecs which we benchmark.

Note that the `Bit-Swap' method, presented by \citet{kingma2019} has a similar
aim to HiLLoC, namely scaling up the method described in \Cref{chap:plc}.
We describe the central idea of Bit-Swap in \Cref{sec:bit-swap}, and where
relevant we compare our approach to it, describing the trade-offs where they
exist. We recommend reading at least \Cref{chap:ans,chap:plc} for background
before trying to understand the material in this chapter.

\begin{figure}[t]
\centering
\includegraphics[width=\textwidth]{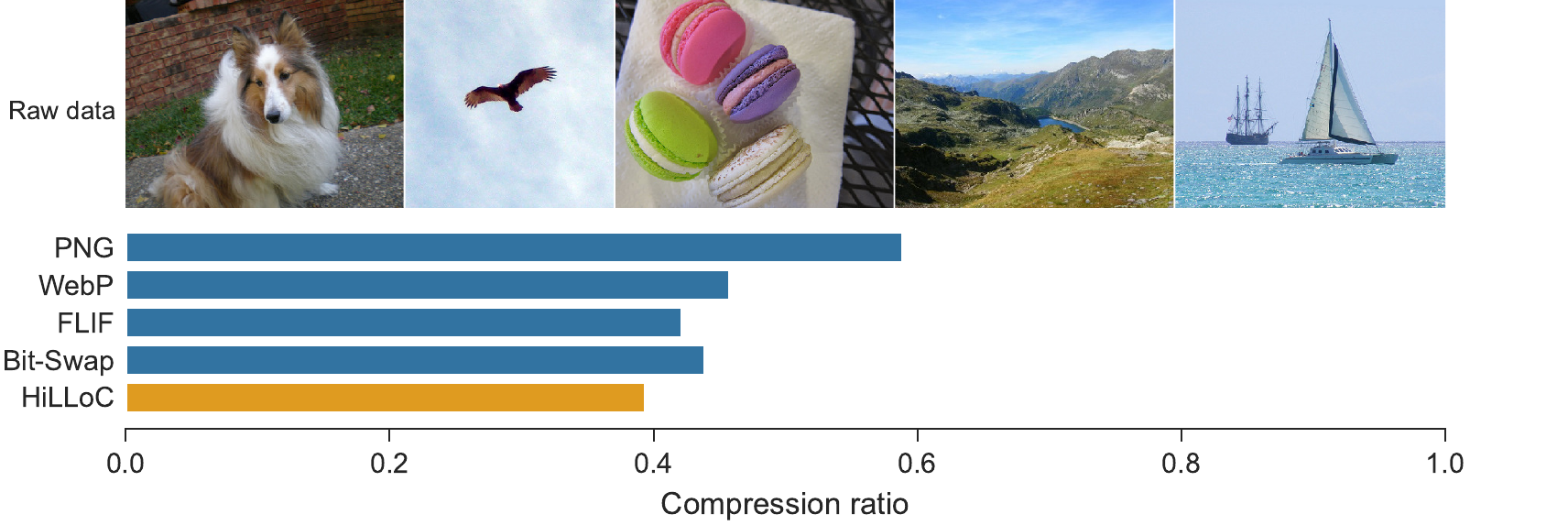}
\caption{
A selection of images from the ImageNet dataset and the compression rates
achieved on the dataset by PNG, WebP, FLIF, Bit-Swap and the HiLLoC codec (with
ResNet VAE) presented in this chapter.}
\label{fig:test_images}
\end{figure}

\section{Scaling up BB-ANS}\label{sec:contributions}
\subsection{Fully convolutional models}
When all of the layers in the generative and recognition networks of a VAE are
either convolutional or element-wise functions (i.e. the VAE has no densely
connected layers), then it is possible to evaluate the recognition network on
images of any height and width, and similarly to pass latents of any height and
width through the generative network to generate an image. Thus, such a VAE can
be used as a (probabilistic) model for images of any size.

We exploit this fact, and show empirically in \Cref{sec:hilloc-experiments}
that, surprisingly, a fully convolutional VAE trained on 32 $\times$ 32 images
can perform well (in the sense of having a high ELBO) as a model for 64
$\times$ 64 images as well as far larger images.  This, in turn, corresponds to
a good compression rate, and we implement lossless compression of arbitrary
sized images by using a VAE in this way. We would expect this result might
extend to deep generative models other than VAEs, i.e. models which use masked
convolutions \citep{germain2015} and flow-based models, although investigating
this is outside of the scope of this work.

\subsection{Reducing one-time overhead}\label{sec:starting}
As mentioned in \Cref{sec:plc-chaining}, if we don't have an `extra message' to
communicate, and we wish to communicate a sequence of data points (modeled as
independent), then we may use a method other than BB-ANS to code the first
data, until enough of a buffer has been built up to generate a latent from
\(Q(z\given x)\), or we may generate a latent \(z\) for the first datapoint,
using any process we like, and encode it with the prior and posterior,
accepting that we will then be commucating redundant information at the
beginning of a chain. The overhead of starting the chain is particularly
significant for hierarchical models, because the information required to sample
from \(Q(z\given x)\) typically scales with the dimensionality of \(z\) and
thus with the latent depth.

In this subsection we discuss two techniques for mitigating this issue. The
first is `Bit-Swap', which was presented in \citet{kingma2019}. Bit-Swap
exploits Markov chain structured latents, by interleaving latent pop and push
steps, and for models with the right structure it can reduce the overhead from
\(O(L)\) to \(O(1)\).

It may sometimes be desirable to use latents which do not have a Markov chain
structure, and for this we use an alternative, simpler method for reducing
overhead (which could also be used to \emph{further} reduce the overhead when
using Bit-Swap), which is simply to use the FLIF codec, which we know to be
reasonably efficient, to code the first elements of the batch.

\subsubsection{Bit-Swap}\label{sec:bit-swap}
The Bit-Swap method \citep{kingma2019} assumes that the generative model has
a hierarchical structure with \(L\) latent layers, illustrated, for a model
with \(L=3\), in \cref{fig:graphicalmodel}. In particular it assumes that the
generative joint distribution and the approximate posterior may be expressed as
\begin{align}
  P(x, z)      &= P(x\given z_1)\left[\prod_{l=1}^{L-1}P(z_l\given
                  z_{l+1})\right]P(z_L)\\
  Q(z\given x) &= Q(z_1\given x)\prod_{l=2}^LQ(z_l\given z_{l-1}).
\end{align}
Given this structure, it is possible to alter the BB-ANS procedure from
\cref{tab:sender_table}, to avoid popping the whole of \(z\) in one go. The
sender can interleave steps, popping one layer at a time and pushing whenever
possible, as shown in \cref{tab:bit-swap}. This technique reduces the initial
bits overhead from \(O(L)\) to \(O(1)\), at the cost of introducing the Markov
restriction on the generative model and approximate posterior (we discuss this
further in \Cref{post_restrictions}).
\begin{table}[ht]
\makebox[\textwidth][c]{
\begin{subtable}{0.52\textwidth}
\begin{tabular}{llll}
Variables & \multicolumn{3}{l}{Operation} \\\midrule
$x$             &$z_1$     & $\leftarrow$  & $Q(\cdot\given x)$  \\\addlinespace
$x,z_1$         &$x$       & $\rightarrow$ & $P(\cdot\given z_1)$\\\addlinespace
$z_1$           &$z_2$     & $\leftarrow$  & $Q(\cdot\given z_1)$\\\addlinespace
$z_1,z_2$       &$z_1$     & $\rightarrow$ & $P(\cdot\given z_2)$\\\addlinespace
$\vdots$        &          & $\vdots$      &                     \\\addlinespace
$z_{L-1},z_L$   &$z_{L-1}$ & $\rightarrow$ & $P(\cdot\given z_L)$\\\addlinespace
$z_L$           &$z_L$     & $\rightarrow$ & $P(\cdot)$          \\\addlinespace
\end{tabular}
\caption{Encoder}
\end{subtable}
\begin{subtable}{0.52\textwidth}
\begin{tabular}{llll}
\multicolumn{3}{l}{Operation} &Variables \\\midrule
$z_L$    &$\leftarrow$ & $P(\cdot)$          &$z_L$         \\\addlinespace
$z_{L-1}$&$\leftarrow$ & $P(\cdot\given z_L)$&$z_{L-1},z_L$ \\\addlinespace
         &$\vdots$      &                     &$\vdots$      \\\addlinespace
$z_1$    &$\leftarrow$ & $P(\cdot\given z_2)$&$z_1,z_2$     \\\addlinespace
$z_2$    &$\rightarrow$  & $Q(\cdot\given z_1)$&$z_1$         \\\addlinespace
$x$      &$\leftarrow$ & $P(\cdot\given z_1)$&$x,z_1$       \\\addlinespace
$z_1$    &$\rightarrow$  & $Q(\cdot\given x)$  &$x$           \\\addlinespace
\end{tabular}
\caption{Decoder}
\end{subtable}
}
\caption{
The Bit-Swap encoder and decoder procedures, in order from the top, for a
model with $L$ layers. For the encoder, the `Variables' column shows variables
known before each operation. For the decoder, that column shows variables known
after each operation.}\label{tab:bit-swap}
\end{table}
\subsubsection{Reducing overheads with FLIF}
As will be explained in Section \ref{sec:discretization}, our dynamic
discretization method precludes the use of Bit-Swap for reducing the one-time
cost of starting a BB-ANS chain, and we also want to be able to use a model
without Markov chain latent structure. We propose instead to use a
significantly simpler method to address the high cost of coding a small number
of samples with BB-ANS: we code the first samples using a different codec. The
purpose of this is to build up a sufficiently large buffer of compressed data
to permit the first stage of the BB-ANS algorithm---to pop a latent sample
from the posterior. In our experiments we use the `Free Lossless Image Format'
\citep[FLIF;][]{sneyers2016} to build up the buffer. We chose this codec
because it was the best performing at the time of writing, but in principal any
lossless codec could be used.

The amount of previously compressed data required to pop a posterior sample
from the ANS stack (and therefore start the BB-ANS chain) is roughly
proportional to the size of the image we wish to compress, since in a fully
convolutional model the dimensionality of the latent space is determined by the
image size.

We can exploit this to obtain a better compression rate than FLIF as quickly as
possible. We do so by partitioning the first images we wish to compress with
HiLLoC into smaller patches. These patches require a smaller data buffer, and
thus we can use the superior HiLLoC coding sooner than if we attempted to
compress full images. We find experimentally that, generally, larger patches
have a better coding rate than smaller patches. Therefore we increase the size
of the image patches being compressed with HiLLoC as more images are compressed
and the size of the data buffer grows, until we finally compress full images
directly once the buffer is sufficiently large.

For our experiments on compressing full ImageNet images, we compress
32$\times$32 patches, then 64$\times$64, then 128$\times$128 before switching
to coding the full size images directly. Note that since our model can compress
images of any shape, we can compress the edge patches which will have different
shape if the patch size does not divide the image dimensions exactly. Using
this technique means that our coding rate improves gradually from the FLIF
coding rate towards the coding rate achieved by HiLLoC on full images. We
compress only 5 full ImageNet images using FLIF before we are able to start
compressing 32$\times$32 patches using HiLLoC.

\subsection{Dynamic discretization}\label{sec:discretization}
It is standard for state of the art latent variable models to use continuous
latent variables.  Since ANS operates over \emph{discrete} probability
distributions, if we wish to use BB-ANS with such models it is necessary to
discretize the latent space so that latent samples can be communicated.  In
\Cref{sec:plc-disc}, we described a \emph{static} discretization scheme for the
latents in a simple VAE with a single layer of continuous latent variables, and
in \Cref{sec:plc-experiments}, we showed that this discretization has a
negligible impact on compression rate.  The addition of multiple layers of
stochastic variables to a VAE has been shown to improve performance
\citep{kingma2016,sonderby2016,maaloe2019,kingma2019}. Motivated by this, we
propose a discretization scheme for hierarchical VAEs with multiple layers of
latent variables.

The discretization described in \Cref{sec:plc-disc} is formed by dividing the
latent space into intervals of equal mass under the prior $p(z)$. For a
hierarchical model, the prior on each layer depends on the previous layers:
\begin{equation}
    p(z_{1:L}) = p(z_L)\prod_{l=1}^{L-1}p(z_l\given z_{l+1:L}).
\end{equation}
It isn't immediately possible to use the simple static scheme from
\Cref{sec:plc-disc}, since the marginals $p(z_1),\ldots,p(z_{L-1})$ are not
known. The Bit-Swap method, described in \Cref{sec:bit-swap}, estimates these
marginals by sampling, creating static bins based on the estimates. They
demonstrate that this approach can work well, see \citet{kingma2019} for
details.  We propose an alternative approach, allowing the discretization to
vary with the context of the latents we are trying to code. We refer to this as
\emph{dynamic discretization}.

In dynamic discretization, instead of discretizing with respect to the
marginals of the prior, we discretize according to the \textit{conditionals} in
the prior, $p(z_l \given z_{l+1:L})$. Specifically, for each latent layer $l$,
we partition each dimension into intervals which have equal probability mass
under the conditional $p(z_l\given z_{l+1:L})$. This directly generalizes the
static scheme from \Cref{sec:plc-disc}.

Dynamic discretization is more straightforward to implement than the method
used with Bit-Swap, because it doesn't require calibrating the discretization
to samples. However it imposes a restriction on model structure, in particular
it requires that posterior inference is done \emph{top-down}. This precludes
the use of the Bit-Swap technique for reducing the size of the initial message
needed to start the bits-back chain.  In Section \ref{post_restrictions} we
contrast the model restriction from dynamic discretization with the bottom-up,
Markov restriction imposed by Bit-Swap itself.

We give further details about the dynamic discretization implementation which
we use in \Cref{app:reparam}.
\begin{figure}[htb]
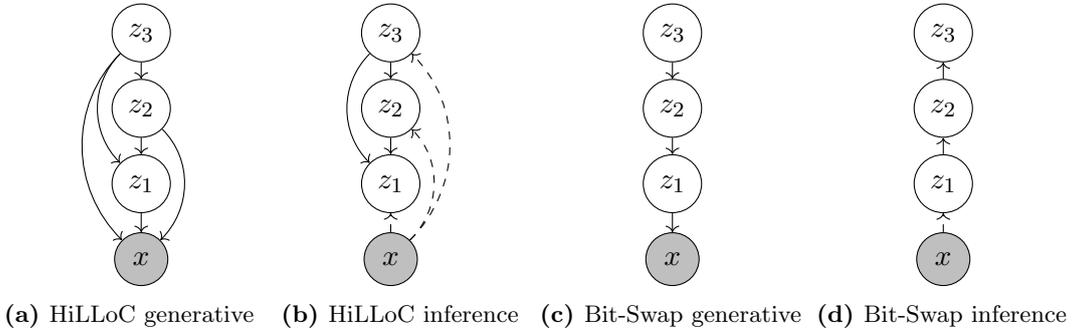

\centering

\makebox[\textwidth][c]{
\begin{subfigure}{.27\textwidth}
\centering
\drawhillocgenerative
\caption{HiLLoC generative}\label{subfig:gen}
\end{subfigure}%

\begin{subfigure}{.27\textwidth}
\centering
\drawhillocinference
\caption{HiLLoC inference}\label{subfig:inf}
\end{subfigure}%

\begin{subfigure}{.27\textwidth}
\centering
\drawbitswapgenerative
\caption{Bit-Swap generative}\label{subfig:bs-gen}
\end{subfigure}%

\begin{subfigure}{.27\textwidth}
\centering
\drawbitswapinference
\caption{Bit-Swap inference}\label{subfig:bs-inf}
\end{subfigure}
}
\caption{
Graphical models representing the generative and inference models with HiLLoC
and Bit-Swap, both using a 3 layer latent hierarchy. The dashed lines indicate
dependence on a fixed observation.}
\label{fig:graphicalmodel}
\end{figure}
\subsubsection{Model restrictions}\label{post_restrictions}
The first stage of BB-ANS encoding is to pop from the posterior,
$z_{1:L}\leftarrow q(\cdot \given x)$. When using dynamic discretization,
popping the layer $z_l$ requires knowledge of the discretization used for $z_l$
and thus of the conditional distribution $p(z_l \given z_{l+1:L})$. This
requires the latents $z_{l+1:L}$ to have already been popped. Because of this,
latents in general must be popped (sampled) in `top-down' order, i.e. $z_L$
first, then $z_{L-1}$ and so on down to $z_1$.

The most general form of posterior for which top-down sampling is tractable is
\begin{equation}
    q(z_{1:L}\given x)=q(z_L\given x)\prod_{l=1}^{L-1}q(z_l\given z_{l+1:L}, x).
\end{equation}
This is illustrated, for a hierarchy of depth 3, in \cref{subfig:inf}.
The Bit-Swap technique \citep{kingma2019} requires that inference be done bottom
up, and that generative and inference models must both be a Markov chain on
$z_1, \ldots,z_L$, and thus cannot use skip connections. These constraints are
illustrated in \cref{subfig:bs-gen,subfig:bs-inf}. Skip connections have been
shown to improve the achievable ELBO value in very deep models
\citep{sonderby2016,maaloe2019}. HiLLoC does not have this constraint, and we
do utilize skip connections in our experiments.

\section{Experimental results}\label{sec:hilloc-experiments}
Using Craystack, we implement HiLLoC with a ResNet VAE
\citep[RVAE;][]{kingma2016}. In all experiments we used an RVAE with 24
stochastic hidden layers.  The RVAE utilizes skip connections, which are
important for effectively training models with such a deep latent hierarchy.
See \Cref{app:rvae_arch} for more details.

We trained the RVAE on the ImageNet 32 training set, then evaluated the RVAE
ELBO and HiLLoC compression rate on the ImageNet 32 test set. To test
generalization, we also evaluated the ELBO and compression rate on the tests
sets of ImageNet64, CIFAR-10 and full size ImageNet. For full size ImageNet, we
used the partitioning method described in \Cref{sec:starting} to compress the
first images. The results are shown in \Cref{tab:hilloc-results}.

For HiLLoC the compression rates are for the entire test set, except for full
ImageNet, where we use 2000 randomly selected images from the test set.

\begin{table}[h]
\caption{
Compression performance of HiLLoC with RVAE compared to other codecs. Rates
measured in bits/dimension (raw data is 8 bits/dimension). For HiLLoC we
display compression rate and theoretical performance (ELBO). All HiLLoC results
are obtained from the \emph{same model}, trained on ImageNet 32.
}\label{tab:hilloc-results}
\centering
\makebox[\textwidth][c]{
\small
\begin{tabular}{llcccc}
                  &                    & ImageNet 32   & ImageNet 64   & Cifar-10      & ImageNet      \\\midrule[\heavyrulewidth]
\emph{Generic}    & PNG                & 6.39          & 5.71          & 5.87          & 4.71          \\
                  & WebP               & 5.29          & 4.64          & 4.61          & 3.66          \\
                  & FLIF               & 4.52          & 4.19          & 4.19          & 3.37          \\
\midrule
\emph{Flow-based} & IDF\tablefootnote{Integer discrete flows, retrieved from
\citet{hoogeboom2019}.}
                                       & 4.18          & 3.90          & 3.34          & -             \\
                  & IDF generalized\tablefootnote{Integer discrete flows trained on ImageNet 32.
                  ImageNet 64 images are split into four 32$\times$32 patches. Retrieved from
                  \citet{hoogeboom2019}.}
                                       & 4.18          & 3.94          & 3.60          & -             \\
                  & LBB\tablefootnote{Local bits back, retrieved from \citet{ho2019}.}
                                       & \textbf{3.88} & \textbf{3.70} & \textbf{3.12} & -             \\
\midrule
\emph{VAE-based}  & Bit-Swap           & 4.50          & -             & 3.82          & 3.51\tablefootnote{
For Bit-Swap, full size ImageNet images were cropped so that their side lengths were multiples of 32.} \\
                  & HiLLoC             & 4.20          & 3.90          & 3.56          & \textbf{3.15} \\
                  & HiLLoC (ELBO)      & (4.18)        & (3.89)        & (3.55)        & (3.14)        \\
\end{tabular}
}
\end{table}

\Cref{tab:hilloc-results} shows that HiLLoC achieves competitive compression
rates on all benchmarks, and is the first deep learning based method to be
evaluated on full size ImageNet images. This significant upscaling relative to
previous work was enabled by the speedups from vectorizing ANS in Craystack. We
anticipate that flow-based and/or autoregressive models may outperform VAEs on
this task in terms of compression rate, possibly with slower
compression/decompression times. Further work is required to investigate
different model classes and architectures.

The fact that HiLLoC with an RVAE can achieve state of the art compression on
full ImageNet relative to the baselines, even under a change of distribution,
is striking.  This provides evidence of its efficacy as a general method for
lossless compression of natural images.  Naïvely, one might expect a
degradation of performance relative to the original test set when changing the
test distribution---even more so when the resolution changes.  However, in the
settings we studied, the \emph{opposite} was true, in that the average
per-pixel negative ELBO (and thus the compressed message length) was
\emph{lower} on all other datasets compared to the ImageNet 32 validation set.

In the case of CIFAR, we conjecture that the reason for this is that its images
are simpler and contain more redundancy than ImageNet. This theory is backed up
by the performance of standard compression algorithms which, as shown in
\Cref{tab:hilloc-results}, also perform better on CIFAR images than they do on
ImageNet 32. We find the compression rate improvement on larger images more
surprising.  We hypothesize that this is because pixels at the edge of an image
are harder to model because they have less context to reduce uncertainty. The
ratio of edge pixels to interior pixels is lower for larger images, thus we
might expect less uncertainty per pixel in a larger image.

To demonstrate the effect of vectorization we timed ANS of single images at
different, fixed, sizes, using a fully vectorized and a fully serial
implementation. The results are shown in Figure \ref{fig:timings} in the
previous chapter, which clearly shows a speed-up of nearly three orders of
magnitude for all image sizes. We find that the run times for encoding and
decoding are roughly linear in the number of pixels, and the time to
compress/decompress an average sized ImageNet image of $500\times374$ pixels
(with vectorized ANS) is around 30s on a desktop computer with 6 CPU cores and
a GTX 1060 GPU. Most of this time is spent on neural net computation,
highlighting the need for more computationally efficient models, and/or more
powerful hardware.

\section{Discussion}
Our experiments demonstrate HiLLoC as a bridge between large scale latent
variable models and compression. To do this we use simple variants of
pre-existing VAE models. Having shown that bits back coding is flexible enough
to compress well with large, complex models, we see plenty of work still to be
done in searching model structures (i.e.\ architecture search), optimizing with
a trade-off between compression rate, encode/decode time and memory usage.
Particularly pertinent for HiLLoC is latent dimensionality, since compute time
and memory usage both scale with this. Since the model must be
stored/transmitted to use HiLLoC, weight compression is also highly relevant.
This is a well-established research area in machine learning
\citep{han2016,ullrich2017}.

Our experiments also demonstrated that one can achieve good performance on a
dataset of large images by training on smaller images. This result is
promising, but future work should be done to discover what the best training
datasets are for coding generic images. One question in particular is whether
results could be improved by \emph{training}, as opposed to just evaluating,
models on larger images and/or images of varying shape and size. We leave this
to future work. Another related direction for improvement is batch compression
of images of different sizes using masking, analogous to how samples of
different length may be processed in batches by recurrent neural nets in
natural language processing applications.

Whilst this work has focused on latent variable models, there is also promise
in applying state of the art fully observed auto-regressive models to lossless
compression. We look forward to future work investigating the performance of
models such as WaveNet \citep{vandenoord2016a} for lossless audio compression
as well as PixelCNN++ \citep{salimans2016} and the state of the art models in
\citet{jun2020} for images. Sampling speed for these models, and thus
decompression, scales with autoregressive sequence length, and can be very
slow. This could be a serious limitation, particularly in common applications
where encoding is performed once but decoding is performed many times. This
effect can be mitigated by using dynamic programming
\citep{lepaine2016,ramachandran2017}, and altering model architecture
\citep{reed2017}, but on parallel architectures sampling/decompression may
still be slower than with VAE models.

On the other hand, fully observed models, as well as the flow based models of
\citet{hoogeboom2019} and \citet{ho2019}, do not require bits back coding, and
therefore do not have to pay the one-off cost of starting a chain.  Therefore
they may be well suited to situations where one or a few i.i.d.\ samples are to
be communicated. Similar to the way that we use FLIF to code the first images
for our experiments, one could initially code images using a fully observed
model then switch to a faster latent variable model once a stack of bits has
been built up.

\section{Conclusion}
In this chapter, we described HiLLoC, an extension of BB-ANS to hierarchical
latent variable models, and show that HiLLoC can perform well with large
models. We open-sourced our implementation, along with the Craystack package
for prototyping lossless compression. We have also explored generalization of
large VAE models, and established that fully convolutional VAEs can generalize
well to other datasets, including images of very different size to those they
were trained on. Finally, we have described a method for compressing images of
arbitrary size with HiLLoC, achieving a compression rate superior to the best
available codecs on ImageNet images.

%% file: Conclusions.tex
\chapter{General Conclusions}\label{chap:conclusion}

The research on which this thesis is based had two aims. The first, more
direct, aim was to develop methods for lossless compression using latent
variable models, and show that they could perform well. The second was broadly
to demonstrate that lossless compression is an interesting and potentially
useful application of recently developed generative modelling techniques, and
that compression ideas can be prototyped by machine learning practitioners
without too much difficulty, particularly using ANS.

Towards the first aim, we have demonstrated that it is practically possible to
implement lossless compression using large scale latent variable models, and in
particular that compression rates very close to the model negative ELBO can be
achieved in large-scale VAEs, on full size images from the ImageNet dataset. We
have also, towards the second aim, presented what we hope is an accessible
introduction to ANS, and easy-to-use software for prototyping ANS-based
compression.

It is too early to say how significant our impact has been in shifting
attention towards lossless compression among machine learning practitioners.
However, there has already a strong response from the community: whilst we're
not aware of any work prior to ours connecting deep generative models to ANS,
since the publication of \citet{townsend2019} there have been at least four
publications by other machine learning researchers which use ANS and deep
generative models to demonstrate new lossless compression ideas
\citep{kingma2019,ho2019,hoogeboom2019,vandenberg2020}. Perhaps more
significantly, as a result of our work, ANS and BB-ANS now form part of the
popular `Deep Unsupervised Learning' course at UC Berkeley. We hope that the
ideas, experimental results and software presented in this thesis can form a
strong basis for future research applying deep learning to lossless
compression.\looseness=-1

%% file: Appendices.tex
\phantomsection
\addcontentsline{toc}{chapter}{Appendices}

% The \appendix command resets the chapter counter, and changes the chapter
% numbering scheme to capital letters.
%\chapter{Appendices}
\appendix
\chapter{Example ANS implementation}\label{app:ans-example}
The following is a complete implementation, in pure Python, of the range
variant of asymmetric numeral systems (rANS), using the same variable names as
in \Cref{chap:ans}. A cons list (implemented as a binary tuple tree) is used as
the stack data structure for \(t\), and the methods \texttt{flatten\_stack} and
\texttt{unflatten\_stack} convert between the stack representation and a Python
list. For more detail, see
\jurl{github.com/j-towns/ans-notes/blob/master/rans.py}.
\begin{singlespace}
\begin{lstlisting}[frame=single,columns=fixed,caption=Scalar rANS core]
s_prec = 64
t_prec = 32
t_mask = (1 << t_prec) - 1
s_min  = 1 << s_prec - t_prec

m_init = s_min, ()  # Shortest possible message

def rans(model):
    f, g, p_prec = model
    def push(m, x):
        s, t = m
        c, p = g(x)
        while s >= p << s_prec - p_prec:
            s, t = s >> t_prec, (t, s & t_mask)
        s = (s // p << p_prec) + s % p + c
        return s, t

    def pop(m):
        s, t = m
        s_bar = s & ((1 << p_prec) - 1)
        x, (c, p) = f(s_bar)
        s = p * (s >> p_prec) + s_bar - c
        while s < s_min:
            t, t_top = t
            s = (s << t_prec) + t_top
        return (s, t), x
    return push, pop

def flatten_stack(t):
    flat = []
    while t:
        t_top, t = t
        flat.append(t_top)
    return flat

def unflatten_stack(flat):
    t = ()
    for t_top in reversed(flat):
        t = t_top, t
    return t
\end{lstlisting}
\end{singlespace}

\begin{singlespace}
\begin{lstlisting}[frame=single,columns=fixed,caption=Scalar rANS example
                   usage,showstringspaces=false]
import math

log = math.log2

# We encode some data using the example model from
# Chapter 2 and verify the inequality in eq. (2.19).

# First setup the model
p_prec = 3

# Probability weights, must sum to 2 ** p_prec
ps = {'a': 1,
      'b': 2,
      'c': 3,
      'd': 2}

# Cumulative probabilities
cs = {'a': 0,
      'b': 1,
      'c': 3,
      'd': 6}

# Backwards mapping
s_bar_to_x = {0: 'a',
              1: 'b', 2: 'b',
              3: 'c', 4: 'c', 5: 'c',
              6: 'd', 7: 'd'}

def f(s_bar):
    x = s_bar_to_x[s_bar]
    c, p = cs[x], ps[x]
    return x, (c, p)

def g(x):
    return cs[x], ps[x]

model = f, g, p_prec

push, pop = rans(model)

# Some data to compress
xs = ['a', 'b', 'b', 'c', 'b', 'c', 'd', 'c', 'c']

# Compute h(xs):
h = sum(map(lambda x: log(2 ** p_prec / ps[x]), xs))
print('Information content of sequence: '
      'h(xs) = {:.2f} bits.'.format(h))
print()

# Initialize the message
m = m_init

# Encode the data
for x in xs:
    m = push(m, x)

# Verify the inequality in eq (20)
eps = log(1 / (1 - 2 ** -(s_prec - p_prec - t_prec)))
print('eps = {:.2e}'.format(eps))
print()

s, t = m
lhs = (log(s) + t_prec * len(flatten_stack(t))
       - log(s_min))
rhs = h + len(xs) * eps
print('Eq (20) inequality, rhs - lhs == {:.2e}'
      .format(rhs - lhs))
print()

# Decode the message, check that the decoded data
# matches original
xs_decoded = []
for _ in range(len(xs)):
    m, x = pop(m)
    xs_decoded.append(x)

xs_decoded = reversed(xs_decoded)

for x_orig, x_new in zip(xs, xs_decoded):
    assert x_orig == x_new

# Check that the message has been returned to its
# original state
assert m == m_init
print('Decode successful!')
\end{lstlisting}
\end{singlespace}
\chapter{Reparameterizing discretized latents in hierarchical VAEs}\label{app:reparam}
After discretizing the latent space, the latent variable at layer $l$ can be
treated as simply an index $i_l$ labeling the interval into which \(z_l\)
falls. We introduce the following notation for pushing and popping according to
a discretized version of the posterior:
\begin{equation}
    i_l \leftrightarrow Q_l(\cdot \given i_{l+1:L}, x),
\end{equation}
where $Q_l(\cdot \given i_{l+1:L}, x)$ is the distribution over the intervals
of the discretized latent space for $z_l$, with interval masses equal to their
probability under $q(z_l\given \tilde{z}_{l+1:L}, x)$. The discretization is
created by splitting the latent space into equal mass intervals under
$p(z_l\given \tilde{z}_{l+1:L})$. We use $\tilde{z}$ to denote the centred
values that can be reconstructed from the indices $i_l$. To be precise,
$\tilde{z}_l(i_l)$ is set to the median value within the interval indexed by
$i_l$, under the prior.  Note that $Q_l$ has an implicit dependence on the
previous prior distributions $p(z_k|z_{k+1:L})$ for $k\geq l$, as these prior
distributions are required to calculate $\tilde{z}_{l+1:L}$ and the
discretization of the latent space. Also note that interval construction is
\emph{implicit}---we never have to do any computation over the set of all
intervals, or store that set in memory, everything we need is computed
efficiently only when required.

Since we discretize each latent layer into intervals of equal mass under the
prior, the prior distribution over the indices $i_l$ reduces to a uniform
distribution over the interval indices, $U(i_l)$, which is not dependent on
$i_{\neq l}$. This allows us to push/pop the $i_l$ according to the prior in
parallel. The full encoding and decoding procedures for a hierarchical latent
model with the dynamic discretization are shown in \Cref{tab:resnet}. Note
that the operations in the two tables are ordered top to bottom.

\begin{table}[ht]
\makebox[\textwidth][c]{
\begin{subtable}{0.52\textwidth}
\begin{tabular}{llll}
Variables & \multicolumn{3}{l}{Operation} \\\midrule
$x$             &$i_L$ & $\leftarrow$ &  $Q_L(\cdot\given x)$ \\\addlinespace
$x, i_L$        &$i_{L-1}$ & $\leftarrow$ & $Q_{L-1}(\cdot\given i_L, x)$ \\\addlinespace
$\vdots$        & & $\vdots$ & \\\addlinespace
$x, i_{2:L}$    &$i_{1}$ & $\leftarrow$& $Q_{1}(\cdot\given i_{2:L}, x)$ \\\addlinespace
$x, i_{1:L}$    &$x$ &$\rightarrow$ & $p(\cdot\given\tilde{z}_{1:L}(i_{1:L}))$ \\\addlinespace
$i_{1:L}$        &$i_{1:L}$ & $\rightarrow$ & $U(\cdot)$ \\
\end{tabular}
\caption{Encoding}
\end{subtable}
\begin{subtable}{0.52\textwidth}
\begin{tabular}{llll}
\multicolumn{3}{l}{Operation} &Variables \\\midrule
$i_{1:L}$&$\leftarrow$ &$U(\cdot)$ &$i_{1:L}$ \\\addlinespace
$x$ &$\leftarrow$ &$p(\cdot\given\tilde{z}_{1:L}(i_{1:L}))$ &$x, i_{1:L}$  \\\addlinespace
$i_1$ & $\rightarrow$ & $Q_1(\cdot \given i_{2:L}, x)$ &$x,i_{2:L}$ \\\addlinespace
$i_2$ & $\rightarrow$ &$Q_2(\cdot \given i_{3:L}, x)$ &$x,i_{3:L}$\\\addlinespace
&$\vdots$ &        & $\vdots$ \\\addlinespace
$i_L$ & $\rightarrow$ &$Q_L(\cdot \given x)$ &$x$\\
\end{tabular}
\caption{Decoding}
\end{subtable}
}
\caption{
The BB-ANS encoding and decoding operations, in order from the top, for a
hierarchical latent model with $L$ layers. The $Q_l$ are posterior
distributions over the indices $i_l$ of the discretized latent space for the
$l$th latent, $z_l$. The discretization for the $l$th latent is created such
that the intervals have equal mass under the prior.}\label{tab:resnet}
\end{table}

\chapter{The ResNet VAE architecture}\label{app:rvae_arch}
A full description of the RVAE architecture is given in \citet{kingma2016}, and
a full implementation can be found in our repository
\url{https://github.com/hilloc-submission/hilloc}, but we give a short
description below.

The RVAE is a hierarchical latent variable model, trained by maximization of
the usual evidence lower bound (ELBO) on the log-likelihood:
\begin{equation}
    \log p(x) \geq \mathbb{E}_{q(z\given x)}\left[\log\frac{b(x, z)}{q(z\given
    x)}\right].
\end{equation}
We use $L$ to denote the depth of the latent hierarchy, and label the latent
layers $z_{1:L}$.  There are skip connections in both the generative model,
$p(x, z_{1:L})$, and the inference model, $q(z_{1:L} \given x)$. Due to our
requirement of using dynamic discretization, we use a top-down inference model
\footnote{Note that in \cite{kingma2016}, this is referred to as `bidirectional
inference'.}. This means that we can factorize \(p\) and \(q\)
\begin{align}
p(x, z_{1:L})&=p(x \given z_{1:L})p(z_L) \prod_{l=1}^{L-1}p(z_l \given z_{l+1:L}) \\
q(z_{1:L}\given x)&=q(z_L\given x)\prod_{l=1}^{L-1}q(z_l\given z_{l+1:L}, x)
\end{align}
and express the ELBO as
\begin{align}
\log p(x) \geq ~ & \mathbb{E}_{q(z_{1:L}\given x)}\left[\log p(x \given
                 z_{1:L})\right] - \kl{q(z_L \given x)}{p(z_L)} \\
&- \sum_{l=1}^{L-1} \mathbb{E}_{q(z_{l+1:L} \given{x})}\left[\kl{q(z_l \given
z_{l+1:L}, x)}{p(z_l \given z_{l+1:L})}\right].
\end{align}
Where $D_{\text{KL}}$ is the KL divergence. As in \citet{kingma2016}, the KL
divergence terms are individually clamped by $\max(D_{\text{KL}}, \lambda)$,
where $\lambda$ is some constant. This is an optimization technique known as
\textit{free bits}, and aims to prevent latent layers in the hierarchy
collapsing to the prior.

Each layer in the hierarchy consists of a ResNet block with two sets of
activations. One set of activations are calculated bottom-up (in the direction
of $x$ to $z_L$), and the other are calculated top-down. The bottom-up
activations are used only within $q(z_{1:L} \given{x})$, whereas the top-down
activations are used by both $q(z_{1:L} \given{x})$ and $p(x, z_{1:L})$. Every
conditional distribution on a latent $z_l$ is parameterized as a diagonal
Gaussian distribution, with mean and covariance a function of the activations
within the ResNet block, and the conditional distribution on $x$ is
parameterized by a discretized logistic distribution. Given activations for
previous ResNet blocks, the activations at the following ResNet block are a
combination of stochastic and deterministic features of the previous latent
layer, as well as from skip connections directly passing the previous
activations. The features are calculated by convolutions.

Note also that all latent layers are the same shape. Since we retained the
default hyperparameters from the original implementation, each latent layer has
32 channels and spatial dimensions half those of the input (e.g.  $\frac{h}{2}
\times \frac{w}{2}$ for input of shape $h \times w$).